\documentclass[journal]{IEEEtran}

\usepackage{graphicx}
\usepackage{amsmath,amssymb} 
\usepackage{color}
\usepackage{epstopdf}
\usepackage{multirow}
\usepackage{authblk}
\usepackage[colorlinks,
           linkcolor=black,
           anchorcolor=black,
           citecolor=black
           ]{hyperref}

\usepackage{ifxetex}
\ifxetex
\usepackage[OT1]{fontenc}
\fi

\usepackage{longtable,multirow,hhline,tabularx,array,makecell,diagbox,colortbl,booktabs}

\usepackage{tikz}
\usepackage{caption}
\usepackage{xcolor}
\usetikzlibrary{mindmap} 
\usetikzlibrary{arrows}
\usetikzlibrary{decorations.pathreplacing}
\usetikzlibrary{trees}
\definecolor{DeepSkyBlue4}{RGB}{0,104,139}
\usetikzlibrary{positioning,shadows,shapes,arrows,calc}

\usepackage{subfigure}

\begin{document}

\title{RhythmNet: End-to-end Heart Rate Estimation from Face via Spatial-temporal Representation}

\author{Xuesong Niu,~\IEEEmembership{Student Member,~IEEE,}
        Shiguang Shan,~\IEEEmembership{Senior Member,~IEEE,}
        Hu Han,~\IEEEmembership{Member,~IEEE,} \\
        Xilin Chen,~\IEEEmembership{Fellow,~IEEE}
        \thanks{This research was supported in part by the National Key R$\&$D Program of China (grant 2017YFA0700800), Natural Science Foundation of China (grants 61672496 and 61702486), External Cooperation Program of Chinese Academy of Sciences (CAS) (grant GJHZ1843). (Corresponding author: Shiguang Shan.)}
        \thanks{X. Niu and X. Chen are with the Key Laboratory of Intelligent Information Processing, Institute of Computing Technology, Chinese Academy of Sciences, Beijing 100190, China, and also with the University of Chinese Academy of Sciences, Beijing 100049, China. E-mail: xuesong.niu@vipl.ict.ac.cn; xlchen@ict.ac.cn.}
        \thanks{H. Han is with the Key Laboratory of Intelligent Information Processing of Chinese Academy of Sciences (CAS), Institute of Computing Technology, CAS, Beijing 100190, China, and also with Peng Cheng Laboratory, Shenzhen, China. E-mail: hanhu@ict.ac.cn.}
        \thanks{S. Shan is with the Key Laboratory of Intelligent Information Processing of Chinese Academy of Sciences (CAS), Institute of Computing Technology, CAS, Beijing 100190, China and University of Chinese Academy of Sciences, Beijing 100049, China, and he is also a member of CAS Center for Excellence in Brain Science and Intelligence Technology. E-mail: sgshan@ict.ac.cn.}
        }
\maketitle

\begin{abstract}
Heart rate (HR) is an important physiological signal that reflects the physical and emotional status of a person. Traditional HR measurements usually rely on contact monitors, which may cause inconvenience and discomfort. Recently, some methods have been proposed for remote HR estimation from face videos; however, most of them focus on well-controlled scenarios, their generalization ability into less-constrained scenarios (e.g., with head movement, and bad illumination) are not known. At the same time, lacking large-scale HR databases has limited the use of deep models for remote HR estimation.
In this paper, we propose an end-to-end RhythmNet for remote HR estimation from the face. In RyhthmNet, we use a spatial-temporal representation encoding the HR signals from multiple ROI volumes as its input. Then the spatial-temporal representations are fed into a convolutional network for HR estimation. We also take into account the relationship of adjacent HR measurements from a video sequence via Gated Recurrent Unit (GRU) and achieves efficient HR measurement. In addition, we build a large-scale multi-modal HR database (named as VIPL-HR\footnote{VIPL-HR is available at: \url{http://vipl.ict.ac.cn/view_database.php?id=15}}), which contains 2,378 visible light videos (VIS) and 752 near-infrared (NIR) videos of 107 subjects. Our VIPL-HR database contains various variations such as head movements, illumination variations, and acquisition device changes, replicating a less-constrained scenario for HR estimation. The proposed approach outperforms the state-of-the-art methods on both the public-domain and our VIPL-HR databases.
\end{abstract}


\IEEEpeerreviewmaketitle

\section{Introduction}

Heart rate (HR) is an important physiological signal that reflects the physical and emotional status of a person; therefore, HR measurement is useful in many applications, such as training aid, health monitoring, and nursing care. Traditional HR measurement usually relies on contact monitors, such as electrocardiograph (ECG) and contact photoplethysmography (cPPG) based sensors, which are inconvenient for the users and thus limit the application scenarios. In recent years, a growing number of studies have been reported on remote HR estimation from videos ~\cite{poh2010non,poh2011advancements,balakrishnan2013detecting,de2013robust,li2014remote,Tulyakov2016Self,wang2017algorithmic}, which allows HR estimation from the skin, i.e., the face area, without contact with a person.

\begin{figure}
\centering
\subfigure[]{
\includegraphics[width=0.45\linewidth]{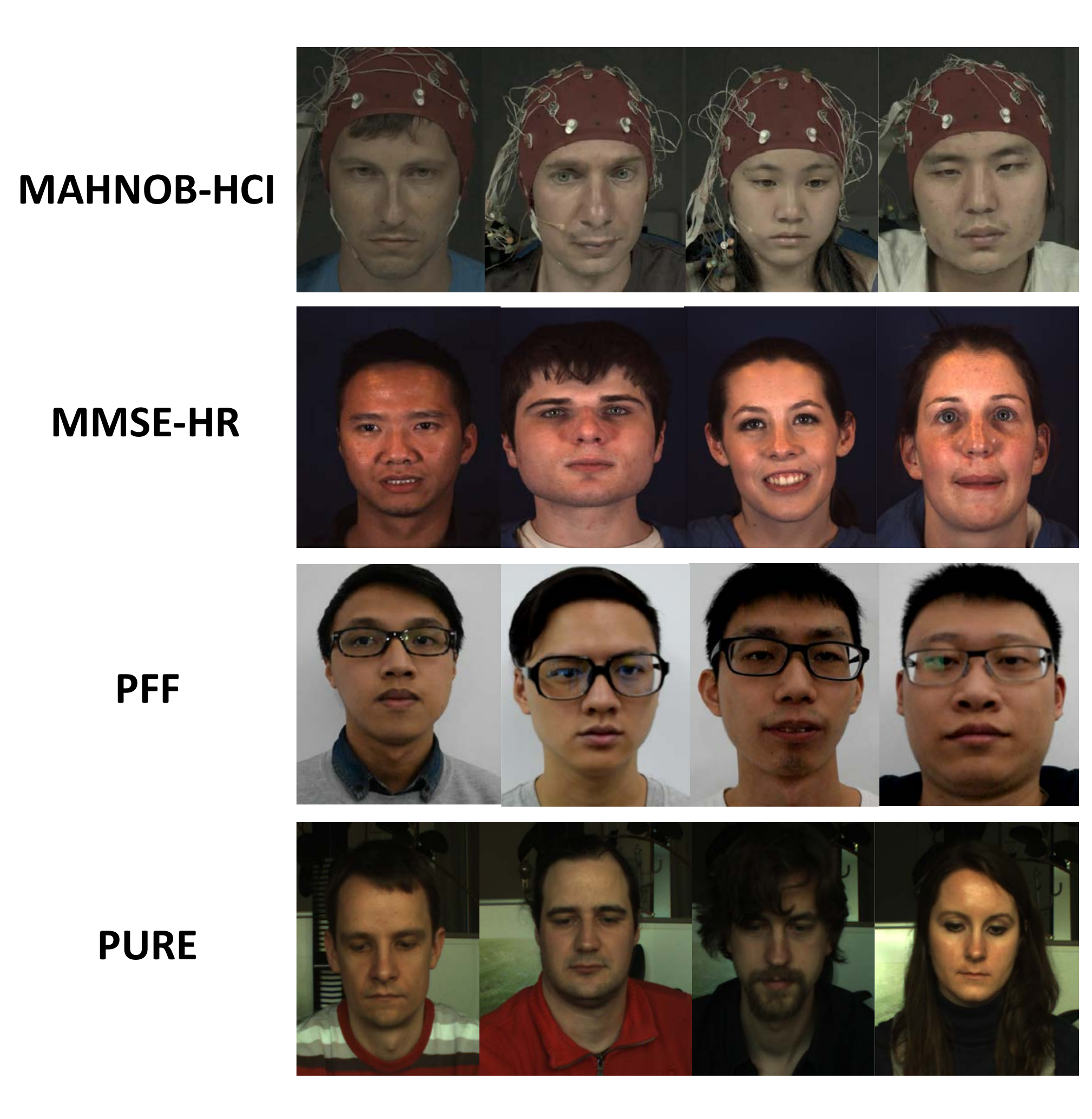}
\label{fig:other_database}
}
\subfigure[]{
\includegraphics[width=0.45\linewidth]{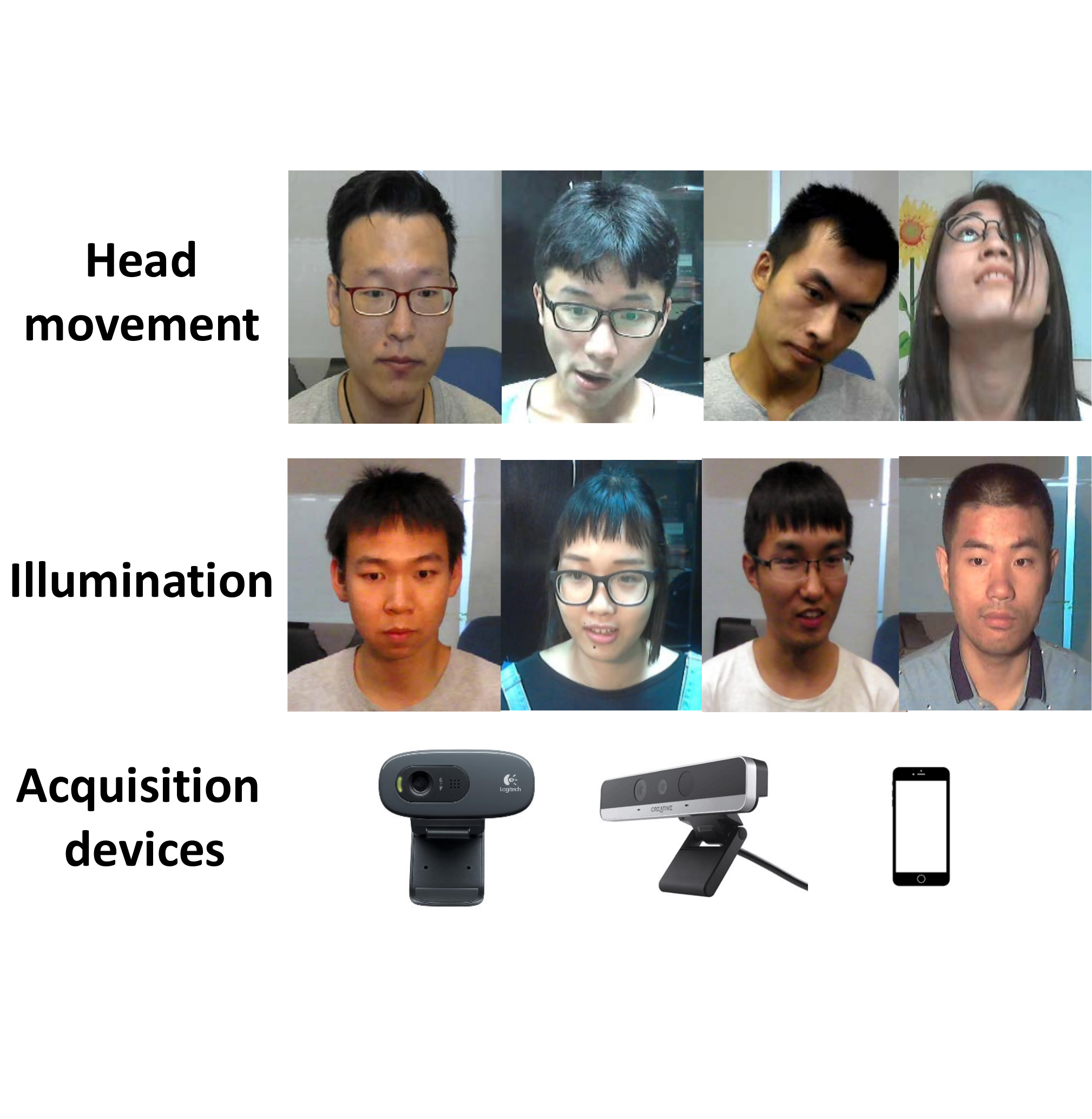}
\label{fig:viplhr_database}
}
\caption{Comparisons between (a) several public domain HR databases (MAHNOB-HCI~\cite{soleymani2012multimodal}, MMSE-HR~\cite{Tulyakov2016Self}, PFF~\cite{hsu2017deep}, PURE~\cite{stricker2014non}) and (b) our VIPL-HR database in terms of illumination condition, head movement, and acquisition device.}
\label{fig:other_vs_VIPLHR}
\end{figure}

The existing video-based HR estimation methods can be grouped into two main categories: remote photoplethysmography (rPPG) based approaches~\cite{poh2010non,poh2011advancements,de2013robust,li2014remote,Tulyakov2016Self,wang2017algorithmic} and the ballistocardiographic (BCG)~\cite{balakrishnan2013detecting,irani2014improved} based approaches.
While rPPG-based approaches aim to extract HR signals from the color changes caused by variations in volume and oxygen saturation of the blood in the vessels due to heart beats, BCG-based approaches aim to extract HR signals from head movements arising from the periodic ejections of blood into the great vessels along with each heartbeat. Most of the existing non-contact HR measurement methods in literature are rPPG-based approaches because it is easier to capture the skin color changes than to get the small head movement in less-constrained scenarios, especially when the subjects perform large-scale head movements. However, head movements and environment conditions, e.g., light conditions, may have great impact on rPPG-based approaches since the amplitude of the HR signal extracted from the color changes of facial skin is too small to be perceived by human eyes.

In order to get stable HR signal, many rPPG-based methods have been proposed. Color space transformation and signal decomposition are commonly used tools for HR signal generation, e.g., using independent component analysis (ICA)~\cite{poh2010non,poh2011advancements} and CHROM~\cite{de2013robust}. These methods reported good results under constrained situations, but may not perform well in the presence of large head motions and lighting variations. Some latter approaches introduced more sophisticated color space transformation and signal decomposition algorithms and obtained improved heart rate measurement accuracy~\cite{Tulyakov2016Self,wang2017algorithmic} under less-constrained situations. However, these methods usually made some certain assumptions w.r.t. the skin reflection (e.g., linear combination assumption~\cite{poh2010non,poh2011advancements}), and the influence by head movement~\cite{de2013robust,wang2017algorithmic}. Such assumptions may not hold in the presence of large head motions and lighting variations. In addition, many prior methods usually perform evaluations on private databases, making it difficult to provide comparisons among different methods. While a few public-domain HR databases are available~\cite{Tulyakov2016Self,soleymani2012multimodal,hsu2017deep,stricker2014non,li2018obf}, the sizes of these databases are quite limited (usually less than 50 subjects) compared with the large scale databases in the other computer vision tasks. Moreover, these databases were usually built in well-controlled scenarios, i,e., with minor illumination and motion variations (see Fig.~\ref{fig:other_database}), which has a big gap with the practical application scenarios.

In this paper, we aim to perform remote HR estimation from face under less-constrained scenarios, where there are large variations in pose and illumination. In one aspect, we aim to build an end-to-end trainable HR estimator leveraging the strong modelling capacity of deep neural networks (DNN) to address the challenges of head movement, illumination variation, and acquisition device difference. In the other aspect, we also hope to provide an effective approach for modelling the relationship between adjacent HR measurement.
To replicate practical scenarios, we also build a large-scale multi-modal HR database (named as VIPL-HR), which consist of 2,378 visible face videos and 752 NIR face videos from 107 subjects. Three different recording devices (RGB-D camera, smart phone, and web-camera) are used for video recording under 9 different situations covering variations in face pose, scale, and illumination (see Fig.~\ref{fig:other_vs_VIPLHR}). We provide evaluations on two widely used public-domain datasets (MAHNOB-HCI~\cite{soleymani2012multimodal} and MMSE-HR~\cite{Tulyakov2016Self}) and our VIPL-HR datasets covering both intra-database and cross-database testing protocols.

This work is an extension of our early work~\cite{niu2018VIPL-HR}. Compared with our previous work, the major extensions in this work are three-fold: i) a detailed review of the published methods for remote HR estimation, and in-depth analysis of the remaining challenges in remote HR estimation from face; ii) we take into account the correlation between adjacent HR measurements and model it via Gated Recurrent Unit (GRU) to achieve robust HR estimation; and iii) extensive evaluations are provided on multiple databases covering a number of aspects of intra-database testing, inter-database testing, visible light imaging, near-infrared imaging, video compression, etc. The experimental results show that the proposed approach outperforms the state-of-the-art algorithms by a large margin.

The rest of this paper is organized as follows. We review the related work of remote HR estimation in Section~\ref{related_work}. The large-scale VIPL-HR database we built is introduced in Section~\ref{VIPL_HR_database}. The proposed end-to-end approach for remote HR estimation is detailed in Section~\ref{Proposed_Approach}. Experimental evaluations and analysis are provided in Section~\ref{experiments}. Finally, we conclude this work in Section~\ref{conculsion}.

\setlength{\tabcolsep}{2pt}
\begin{table*}
\begin{center}
\caption{A brief summary of the existing rPPG based remote HR estimation methods.}
\label{table:existing_method}
\begin{tabular}{lcccccc}
\toprule
\multirow{2}{*}{\textbf{Publication}} & \multirow{2}{*}{\textbf{Camera}} & \multirow{2}{*}{\textbf{Input singal}} & \textbf{HR temporal} & \multirow{2}{*}{\textbf{HR estimation}} & \textbf{Results}\\
& & & \textbf{signal extraction}& & \textbf{(RMSE in bpm)}\\
\toprule
\multirow{2}{*}{Poh et al.~\cite{poh2010non}}  &\multirow{2}{*}{Webcam} & \multirow{2}{*}{R,G,B} & ICA + bandpass & \multirow{2}{*}{FFT} & MAHNOB-HCI\\
 & & & filter & & 25.9 bpm \\
\hline
\multirow{2}{*}{Poh et al.~\cite{poh2011advancements}}  &\multirow{2}{*}{Webcam} & \multirow{2}{*}{R,G,B} & ICA + bandpass \& & \multirow{2}{*}{Peak detection} & MAHNOB-HCI \\
& & & detrending filter & & 13.6 bpm \\
\hline
\multirow{2}{*}{McDuff et al.~\cite{mcduff2014improvements}}  & Five-band & \multirow{2}{*}{R,G,B,C,O} & ICA + bandpass\& & \multirow{2}{*}{Peak detection} & \multirow{2}{*}{-}\\
& camera & & detrending filter & & \\
\hline
\multirow{2}{*}{Lam et al.~\cite{Lam2015Robust}}  &\multirow{2}{*}{Webcam} & \multirow{2}{*}{R,G,B} & Patch level ICA + bandpass\&  & FFT  & MAHNOB-HCI \\
& & &detrending filter & + majority vote & 8.9 bpm \\
\hline
\multirow{2}{*}{Lewandowska et al.~\cite{lewandowska2011measuring}}  &\multirow{2}{*}{Webcam} & \multirow{2}{*}{R,G,B} & PCA + bandpass & \multirow{2}{*}{FFT} & \multirow{2}{*}{-} \\
& & & filter & &  \\
\hline
\multirow{2}{*}{Tsouri et al.~\cite{tsouri2012constrained}}  &\multirow{2}{*}{Webcam} & \multirow{2}{*}{R,G,B} & constrained ICA + bandpass & \multirow{2}{*}{FFT} & \multirow{2}{*}{-} \\
& & & filter & &  \\
\hline
\multirow{2}{*}{Haan and Jeanne~\cite{de2013robust}}  &\multirow{2}{*}{Webcam} & \multirow{2}{*}{R,G,B} & Color space transformation based & \multirow{2}{*}{FFT} & MAHNOB-HCI\ /\ MMSE-HR\ /\ VIPL-HR\\
& & & on skin model (CHROM feature) & & 6.23 bpm\ /\ 11.37 bpm\ /\ 16.9 bpm \\
\hline
\multirow{2}{*}{Wang et al.~\cite{wang2015exploiting}}  &\multirow{2}{*}{Webcam} & \multirow{2}{*}{R,G,B} & Pixel level CHROM  & \multirow{2}{*}{FFT} &  \multirow{2}{*}{-} \\
& & & feature + PCA & &  \\
\hline
\multirow{2}{*}{Feng et al.~\cite{feng2015motion}}  &\multirow{2}{*}{Webcam} & \multirow{2}{*}{R,G} & Adaptive Green/Red  & \multirow{2}{*}{FFT} & \multirow{2}{*}{-} \\
& & & difference & &  \\
\hline
\multirow{2}{*}{Haan and Leest~\cite{de2014improved}} &\multirow{2}{*}{Webcam} & \multirow{2}{*}{R,G,B} & Color space transformation based & \multirow{2}{*}{FFT} & \multirow{2}{*}{-} \\
& & & on skin-tone and camera properties & &  \\
\hline
\multirow{2}{*}{Wang et al.~\cite{wang2017algorithmic}}  &\multirow{2}{*}{Webcam} & \multirow{2}{*}{R,G,B} & Color space transformation based & \multirow{2}{*}{FFT} & VIPL-HR \\
& & & on skin model & &  17.2 bpm \\
\hline
\multirow{2}{*}{Kumar et al.~\cite{kumar2015}}  &\multirow{2}{*}{Webcam} & \multirow{2}{*}{G} & Patch level signals combination  & \multirow{2}{*}{FFT} & \multirow{2}{*}{-} \\
& & & using weighted average & &  \\
\hline
\multirow{2}{*}{Wang et al.~\cite{wang2017amplitude}}  &\multirow{2}{*}{Webcam} & \multirow{2}{*}{R,G,B} & Using the prior knowledge  & \multirow{2}{*}{FFT} & \multirow{2}{*}{-} \\
& & & of heart rate to remove noise & &  \\
\hline
\multirow{2}{*}{Niu et al.~\cite{niucontinuous}}  &\multirow{2}{*}{Webcam} & \multirow{2}{*}{R,G,B} & \multirow{2}{*}{Patch level CHROM feature} & FFT & MAHNOB-HCI \\
& & & &+ distribution learning & 8.72 bpm \\
\hline
\multirow{2}{*}{Tulyakov et al.~\cite{Tulyakov2016Self}}  &\multirow{2}{*}{Webcam} & \multirow{2}{*}{R,G,B} & CHROM features + self-adaptive & \multirow{2}{*}{FFT} & MAHNOB-HCI\ /\ MMSE-HR\ /\ VIPL-HR \\
& & & matrix completion & & 6.23 bpm\ /\ 11.37 bpm\ /\ 21.0 bpm\\
\hline
\multirow{2}{*}{Wang et al.~\cite{wang2016novel}}  &\multirow{2}{*}{Webcam} & \multirow{2}{*}{R,G,B} & Color space transformation based & \multirow{2}{*}{FFT} & \multirow{2}{*}{-} \\
& & & on spatial subspace of skin-pixels & &  \\
\hline
\multirow{2}{*}{Hsu et al.~\cite{hsu2014learning}}  &\multirow{2}{*}{Webcam} & \multirow{2}{*}{R,G,B} & \multirow{2}{*}{ICA} & \multirow{2}{*}{FFT+SVR} & \multirow{2}{*}{-} \\
& & &  & &  \\
\hline
\multirow{2}{*}{Hsu et al.~\cite{hsu2017deep}}  &\multirow{2}{*}{Webcam} & \multirow{2}{*}{R,G,B} & CHROM feature + detrending   & STFT & MAHNOB-HCI \\
& & & filter + {2nd order difference}  & + VGG model & 4.27 bpm \\
\hline
\multirow{2}{*}{Chen et al.~\cite{chen2018deepphys}}  &\multirow{2}{*}{Webcam} & \multirow{2}{*}{R,G,B} & Convolutional Network with & \multirow{2}{*}{FFT} & MAHNOB-HCI (MAE) \\
& & & attention module &  & 4.57 bpm  \\
\hline
\multirow{2}{*}{Li et al.~\cite{li2014remote}} & \multirow{2}{*}{Webcam} & \multirow{2}{*}{G} & Background noise removing  & \multirow{2}{*}{FFT} & MAHNOB-HCI\ /\ MMSE-HR\\
& & & + bandpass \& detrending filter& &  7.62 bpm / 19.95 bpm\\
\hline
\multirow{2}{*}{Van et al.~\cite{van2015motion}} & NIR  & \multirow{2}{*}{NIR} & Signal modified using prior  & \multirow{2}{*}{FFT} & \multirow{2}{*}{-} \\
& Sensor & & of skin reflectance & &  \\
\hline
\multirow{2}{*}{Chen et al.~\cite{chen2016realsense}}  &\multirow{2}{*}{RealSense} & \multirow{2}{*}{NIR} & \multirow{2}{*}{Global self-similarity filter} & \multirow{2}{*}{FFT} & \multirow{2}{*}{-} \\
&&&&&\\
\hline
\multirow{2}{*}{Proposed} & Webcam/Phone & \multirow{2}{*}{R,G,B/NIR} & \multirow{2}{*}{Spatial-temporal maps}  & Deep regression & MAHNOB-HCI\ /\ MMSE-HR\ /\ VIPL-HR \\
& RealSense & & & model with GRU &  4.00 bpm\ /\ 5.03 bpm/\ 8.14 bpm\\
\bottomrule
\end{tabular}
\end{center}
\end{table*}

\section{Related Work}
\label{related_work}


\begin{figure}
\centering
\includegraphics[width=0.85\linewidth]{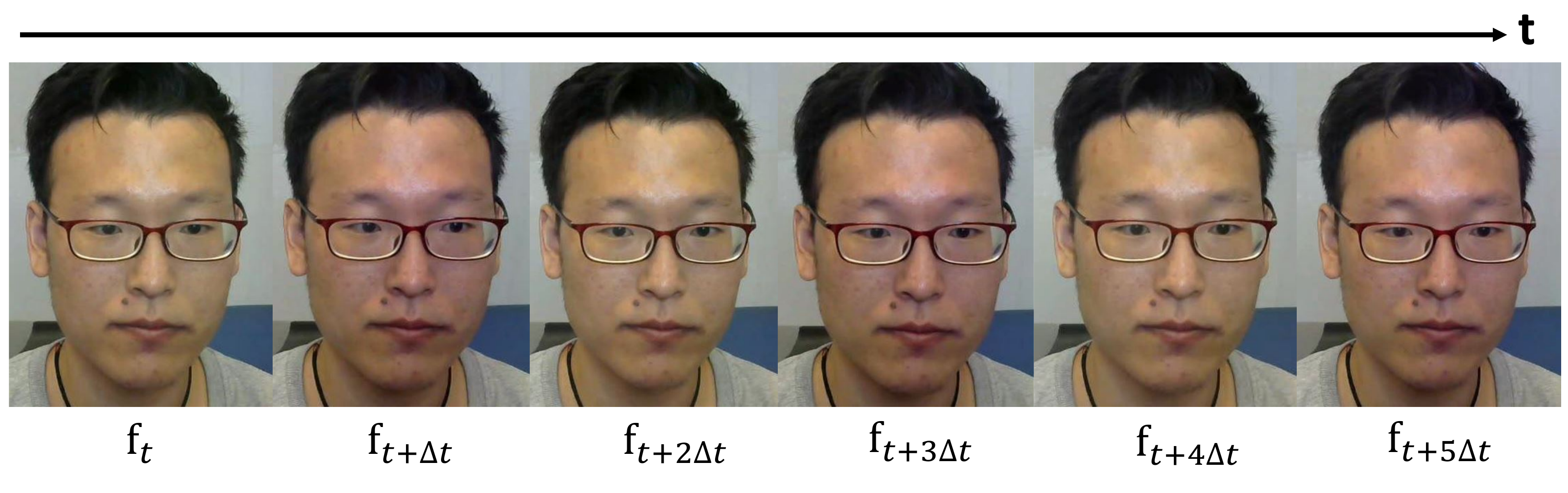}
\caption{Magnified skin color changes due to heartbeat using Eulerian Video Magnification~\cite{wu2012eulerian}. Best viewed in color.}
\label{fig:skin_color_change}
\end{figure}

\subsection{Remote Heart Rate Estimation from Face using rPPG}
\label{related_work_method}
The principle of the rPPG-based HR estimation is based on the fact that optical absorption of skin varies periodically with the blood volume pulse (BVP). Human skin can be regarded as a three-layer model, subcutis, dermis and epidermis. The light is usually absorbed by the chromophores in the skin. The major chromophores of human skin are hemoglobin in the microvasculars of dermis and subcutis layers and melanin in the epidermis layer. The changes of hemoglobin content during a cardiac cycle would cause tiny color changes of the skin\footnote{We provide a video (MagnificatedSkinColorChange.avi) of such magnified skin color changes in \url{https://drive.google.com/file/d/1_sjPOaXP8lFLQkGotJIPsDuU0mRSxegv/view?usp=sharing}} (see Fig.~\ref{fig:skin_color_change}). Although the color changes are invisible to human eyes, they can be captured by commodity RGB sensors and used for computing the HR~\cite{verkruysse2008remote}, and a number of approaches have been reported to use such information for HR estimation.


Although it is possible to perform rPPG based HR measurement, it is still a challenge task since the amplitude of the light absorption variation w.r.t. BVP is minor. 
In addition, these signals can be easily affected by head movements, illumination conditions and noise introduced by the recording devices, making it hard to build a robust HR measurement system under less-constrained situations. A database considering the influences of these factors are also lacking to the community.

Blind signal separation (BSS) methods were introduced for remote HR estimation in~\cite{poh2010non,poh2011advancements}, in which independent component analysis (ICA) was applied to the temporal RGB color signal sequences to find the dominant component related to HR. In~\cite{mcduff2014improvements}, ICA was used for color signals captured by a five-band camera instead of an RGB camera. Different from these approaches which applied ICA to the whole region of interest (ROI), Lam and Kuno applied ICA in a patch level for HR estimation~\cite{Lam2015Robust}. Besides ICA, principal components analysis (PCA)~\cite{lewandowska2011measuring} and constrained ICA (cICA)~\cite{tsouri2012constrained} were also used to extract the purified HR signal from the input RGB video sequence. The BSS-based HR estimation approaches usually make certain assumptions about the HR signals. When these assumptions hold, BSS-based methods can effectively improve the signal-to-noise rate (SNR) of the HR signals. However, the assumptions may not hold under less-constrained scenarios, e.g., with large head movement, poor illumination, diverse sensors, etc. As a result, the HR estimation accuracy of BSS-based methods could degrade severely.

Since the rPPG-based methods is highly related to the skin, it is important to leverage the prior knowledge of skin to improve the robustness. An optical skin model considering head movement was proposed in~\cite{de2013robust}, and used for calculating a chrominance feature representation to reduce the influence of head movement to HR estimation. In a later work of~\cite{wang2015exploiting}, pixel-wise chrominance features were computed and used for HR estimation. Instead of modelling the influence of head movement w.r.t. different color channels, Feng et al.~\cite{feng2015motion} treated the human face as a Lambertian radiator and provided a radiance analysis of different motion types. Besides modelling the optical status of skin under head motion, Haan and Leest~\cite{de2014improved} took the camera optical properties into consideration and proposed a linear decomposition of RGB signals to perform HR estimation. A detailed discussion of different skin optical models for rPPG-based HR estimation was presented in~\cite{wang2017algorithmic}, in which a new projection method was also proposed for extracting pulse signals from an input RGB video. The various skin models utilized by individual HR estimation methods lead to improved robustness. However, when it comes to some complicated scenarios, the assumptions used by these models may not hold, and the HR estimation accuracy could drop significantly. Models that can adaptively deal with various variations in practical application scenarios are required to build a robust HR estimation system. Thus, data-driven based feature learning methods may have the ability to obtain such a representation.

After obtaining the signals related to HR, a common operation is to apply Fast Fourious Transform (FFT) to the signals to get the spectral distribution. The peak of the spectral distribution is considered as the HR frequency (in Hz). Directly applying FFT to the HR signals may get a spectral distribution consisting of many spectral noises. Therefore, a few approaches have been proposed to reduce such noises. In~\cite{kumar2015}, Kumar et al. used the frequency characteristics as the weights to combine the HR signals from different ROIs in order to get a better HR signal. Instead of enhancing the HR signals, Wang et al.~\cite{wang2017amplitude} directly adjusted the spectral distribution according to the prior knowledge of the HR signal. Similarly, Niu et al.~\cite{niucontinuous} took the continuous estimation situations into consideration and used preceding estimations to learn an HR distribution and use it to modulate the spectral distribution. While the frequency domain representation is helpful for remote HR estimation, temporal domain information can also be helpful.

All the methods mentioned above mainly focus on improving the hand-crafted pipelines from periodical temporal signal extraction to signal analysis. Their generalization ability can be poor under unconstrained scenarios. At the same time, data-driven based methods are believed to have strong ability to learn more relevant features to a specific task, particularly when there are given enough data. Some recent methods tried to obtain better temporal signals of the HR from a face video. Tulyakov et al.~\cite{Tulyakov2016Self} divided the face into multiple ROI regions to get a matrix of temporal representation and utilized matrix completion to purify the signals related to HR. Wang et al.~\cite{wang2016novel} proposed a subject-dependent approach to compute the HR signal using spatial subspace rotation (SSR) given a complete continuous sequence of a subject's face. Another kind of data-driven methods for remote HR estimation is to learn a HR estimator using the frequency domain representations as input. Hsu et al.~\cite{hsu2014learning} combined the frequency domain features from RGB channels and the ICA components to get a HR signal representation and used support vector regression (SVR) to estimate the HR. Hsu et al.~\cite{hsu2017deep} generated time-frequency maps from the pre-processed green channel signals and used them as input of a VGG-16 model to estimate the HR. Chen et al.~\cite{chen2018deepphys} fed the original RGB face video into a convolutional network with attention mechanism and output the related HR signals. Although the existing data-driven approaches attempted to make use of statistical learning, as opposed to physical model based signal analysis, they failed to build an end-to-end HR estimator. In addition, the feature representations used by these methods remain hand-crafted, which may not be optimum for the HR estimation task.

Besides rPPG-based HR measurement, another kind of remote HR estimation methods is based on the ballistocardiographic (BCG) signals, which is the subtle head motions caused by cardiovascular circulation. Inspired by the Eulerian magnification method~\cite{wu2012eulerian}, Balakrishnan et al. performed facial key points tracking and used PCA to extract the pulse signal from the trajectories of key points~\cite{balakrishnan2013detecting}. Instead of using the PCA component of key points' trajectories to represent HR signal, Irani et al.~\cite{irani2014improved} performed a stochastic search method using discrete cosine transform (DCT) to find the PCA component with the most significant periodicity. Since this kind of methods is based on subtle motion analysis, the voluntary movements by the subjects will significantly influence the HR estimation accuracy, leading to very limited use in practical applications.

Camera sensors can also have impact on the HR estimation accuracy. Besides the widely used commodity RGB sensors, several approaches also studied the possibility of using non-RGB, e.g., near-infra-read (NIR) sensors. Inspired by the PVB pulse-extraction method in~\cite{de2014improved}, Gastel et al. investigated the feasibility of rPPG-based HR estimation in the NIR spectrum and achieved a robust HR estimation result with both NIR and color sensors~\cite{van2015motion}. Chen et al.~\cite{chen2016realsense} directly used the NIR images captured by RealSense\footnote{\url{https://www.intel.com/content/www/us/en/homepage.html}}  and demonstrated the possibility of using NIR images for HR measurement. Instead of using non-color sensors to perform rPPG-based HR estimation, a BCG-based method using depth images was studied in~\cite{yang2017estimating}. Similar to the color sensor based methods, all the remote HR estimation using non-color sensors are still hand-crafted, and make certain assumptions.

The existing remote HR measurement methods are summarized in Table~\ref{table:existing_method}. Despite tremendous progress has been made in remote HR estimation, there are still limitations. First, the existing approaches usually made certain assumptions when building the model, which may limit the application scenarios. Second, most of the existing approaches are designed in a step-by-step hand-crafted way requiring sophisticated knowledge and experiences. Therefore, end-to-end trainable approaches are required to bridge the gap between the capability of remote HR estimation methods and the real application requirements.

{\setlength{\tabcolsep}{2mm}
\begin{table*}
\begin{center}
\caption{A summary of the public-domain databases and our VIPL-HR database for remote HR estimation.}
\label{table:database}
\begin{tabular}{cccccccc}
\toprule
  & \multirow{2}{*}{\textbf{$\#$ Subjects}} & \textbf{Illumination} & \textbf{Head} & \textbf{Sensor} & \multirow{2}{*}{\textbf{$\#$Videos}} \\
  & & \textbf{variation} & \textbf{movement} & \textbf{diversity} & \\
\toprule
MAHNOB-HCI~\cite{soleymani2012multimodal} & 27 &\emph{L}  & \emph{E} & \emph{C} & 527 \\
MMSE-HR~\cite{Tulyakov2016Self}	& 40 &\emph{L} 	& \emph{E} & \emph{C} & 102  \\
PURE~\cite{stricker2014non}  & 10 &\emph{L}	 & \emph{S/SM/T}  & \emph{C} & 60  \\
PFF~\cite{hsu2017deep}	& 13 &\emph{L/D} & \emph{S/SM} & \emph{C} & 104  \\
OBF~\cite{li2018obf} & 106 &\emph{L}  &\emph{S} & \emph{C/N} & 2,120 \\
\hline
\noalign{\smallskip}
VIPL-HR & \textbf{107} & \textbf{L/D/B} & \textbf{S/LM/T} & \emph{\textbf{C/N/P}} & \textbf{3,130} \\
\bottomrule
\end{tabular}
\smallskip
{\fontsize{8pt}\baselineskip\selectfont
\\ L = Lab Environment, D = Dim Environment, B = Bright Environment, E = Expression, \\ S = Stable, SM = Slight Movement, LM = Large Movement, T = Talking, \\ C = Color Camera, N = NIR Camera, P = Smart Phone Frontal Camera}
\end{center}
\end{table*}
}

\subsection{Public Databases for Remote HR Estimation}

Many existing methods reported their performance using private databases, leading to difficulties in performance comparison across individual approaches. Public-domain database for evaluating remote HR measurement methods are quite limited.
MAHNOB-HCI database~\cite{soleymani2012multimodal} was firstly used for remote HR estimation in~\cite{li2014remote}, which consists of 527 videos of 27 subjects, recorded with small head movement and facial expression variation under laboratory illumination.  MMSE-HR~\cite{Tulyakov2016Self} was introduced by Tulyakov et al., which consists of 102 videos of 40 subjects involving various subjects' facial expressions and head movements. These two databases are originally designed for emotion analysis, and the subjects' variousness is mainly limited to facial expression changes. At the same time, all the videos of MAHNOB-HCI and MMSE-HR are compressed, which may cause some damages to the HR signals used for further calculation.

There are also a few public-available databases especially designed for the task of remote HR estimation. Stricker et al. ~\cite{stricker2014non} released the PURE database consisting of 60 videos from 10 subjects, in which all the subjects are asked to perform 6 kinds of movements such as talking or head rotation. Hus et al.~\cite{hsu2017deep} released the PFF database, consisting of 104 videos of 10 subjects, in which only small head movements and illumination variations are involved. These two databases are limited by the number of subjects and recording scenarios, making them not suitable for building a real-world HR estimator. In 2018, Xiaobai et al.~\cite{li2018obf} proposed the OBF database specifically designed for heart rate variability (HRV) analysis, and all the scenarios in this database are well-controlled, making it very easy to measure HR.

We summarize the public-domain databases for remote HR estimation in Table~\ref{table:database}. We can see that all the existing databases are limited in either the number of subjects or the recording scenarios, particularly limited to constrained scenarios. A large-scale database recorded under less-constrained environment is required for studying remote HR estimation methods in practice.

\section{VIPL-HR database}
\label{VIPL_HR_database}

Considering the constrained data acquisition scenarios used by most of the existing databases for remote HR estimation, we built a new dataset, named as VIPL-HR to promote the research of remote HR estimation under less-constrained scenarios, i.e., with head movement, illumination variation, and acquisition device diversity. VIPL-HR contains 3,130 face videos from 107 subjects, which are recorded with three different sensors (webcam, RealSense, and smartphone) under varying illumination and pose variations. In this section, we provide the details in building our VIPL-HR database covering: i) device setup and data collection, ii) face video compression, and iii) database statistics.

\subsection{Device Setup and Data Collection}

We design our data collection procedure with the objective in mind that face video recording conditions should cover diversities of environmental illumination, subject pose, acquisition sensor, and HR distribution in order to replicate daily application scenarios.

\begin{figure}
\centering
\includegraphics[width=0.8\linewidth]{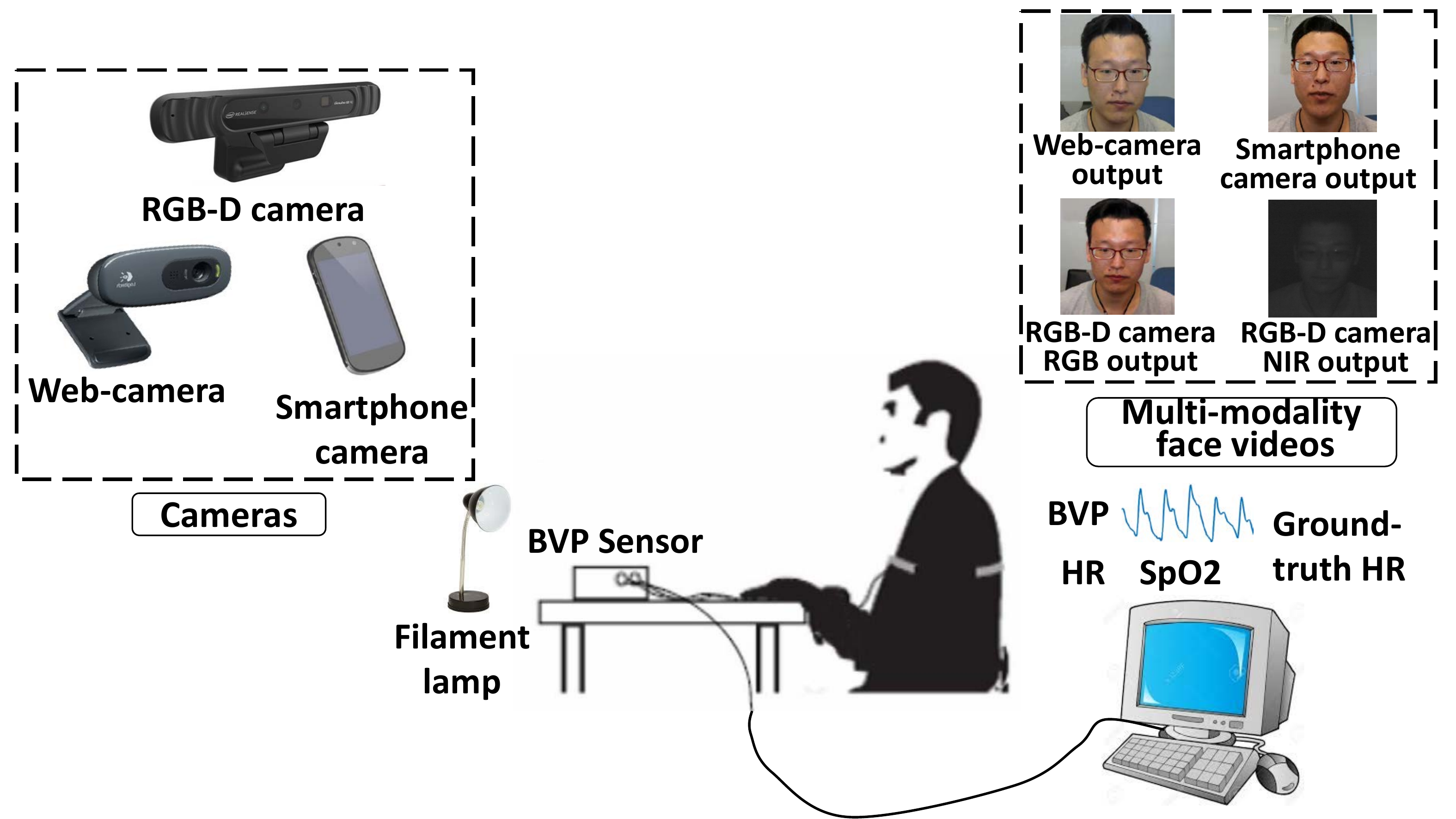}
\caption{Devices and setup used in collecting VIPL-HR.}
\label{fig:setup}
\end{figure}

The recording environmental setup is illustrated in Figure~\ref{fig:setup}. In order to cover diverse illumination conditions, we utilize a filament lamp placed in front of the subject as well as the ceiling lamp. Three different illumination scenarios are conducted, i.e., both lamps are turned on, only the ceiling lamp is turned on, and both lamps are turned off. In order to cover diverse pose variation of the subject, we first ask the subjects to sit casually in front of the cameras, then encourage them to perform daily activities such as talking and looking around. In order to cover a wider range of HR in our VIPL-HR database, we ask each subject to do some exercises and then came back for database recording. At the same time, different distances between the subject and the cameras are also considered.

Sensor diversity is also considered when building the database. We choose three different cameras, i.e., Logitech C310 webcam, Intel Realsense F200 RGB-D camera, and Huawei P9 smartphone (with its frontal camera). Logitech C310 and RealSense F200 are used for visible face video recording with different video resolution, and RealSense F200 is also used for NIR face videos recording to investigate the possibility of remote HR estimation under dark lighting conditions. Since smartphones have become an indispensable part of our daily lives, we also record the face videos when the smartphone is placed in front of the subjects. At the same time, the subjects are also asked to hold the smartphone by themselves to record videos like a video chat scenario. All the related physiological signals, including HR, SpO2, and BVP signals, are synchronously recorded with a CONTEC CMS60C BVP sensor. The details of these recording scenarios can be found in Table~\ref{table:record_situation} and the details of the device specifications can be found in Table~\ref{table:devices}

\setlength{\tabcolsep}{4pt}
\begin{table*}
\begin{center}
\caption{Specifications of individual recording devices used in our VIPL-HR database.}
\label{table:devices}
\begin{tabular}{cccc}
\toprule
\textbf{Device} & \textbf{Specification} & \textbf{Setting}  & \textbf{Output}\\
\toprule
Computer & Lenovo ThinkCentre & Windows 10 OS & N/A\\
\hline
\multirow{2}{*}{Color camera} &  \multirow{2}{*}{Logitech C310} & 25fps$^{*}$ & \multirow{2}{*}{Color videos}\\
&& 960$\times$720 color camera &\\
\hline
\multirow{2}{*}{RGB-D camera} & \multirow{2}{*}{RealSense F200} & 30fps$^{*}$, 640$\times$480 NIR camera & Color videos \\
 & & 1920$\times$1080 color camera, & NIR videos \\
\hline
\multirow{2}{*}{Smart phone}	& HUAWEI P9 & 30fps,  & \multirow{2}{*}{Color videos}\\
 &  frontal camera & 1920$\times$1080 color camera & \\
\hline
\multirow{2}{*}{BVP recoder} & \multirow{2}{*}{CONTEC CMS60C} & \multirow{2}{*}{N/A} & HR, SpO2, \\
& & & and BVP signals \\
\hline
Filament lamp & N/A & 50Hz & N/A \\
\bottomrule
\end{tabular}
\smallskip
{\\$^{*}$ The frame rates of Logitech C310 and RealSense F200 are not very stable, \\and the time steps corresponding to each frame are recorded.}
\end{center}
\end{table*}

\setlength{\tabcolsep}{2pt}
\begin{table}
\begin{center}
\caption{Details of the nine recording situations in the VIPL-HR database.}
\label{table:record_situation}
\begin{tabular}{cccccc}
\toprule
\multirow{2}{*}{\textbf{Scenario}} & \textbf{Head} & \multirow{2}{*}{\textbf{Illumination}} & \multirow{2}{*}{\textbf{Distance}} & \multirow{2}{*}{\textbf{Exercise}} & \textbf{Phone}\\
 & \textbf{movement} & & & & \textbf{recording method} \\
\toprule
1 & S & L & 1m & No & Fixed \\
2 & LM & L & 1m & No & Fixed \\
3 & T & L & 1m & No & Fixed\\
4 & S & B & 1m & No & Fixed\\
5 & S & D& 1m & No & Fixed \\
6 & S & L & 1.5m & No & Fixed\\
7 & S & L & 1m & Yes & Fixed \\
8 & S & L & - & No & Hand-held \\
9 & LM & L & - & No & Hand-held \\
\bottomrule
\noalign{\smallskip}
\end{tabular}
{
\\ S = Stable, LM = Large Movement, T = Talking, \\ L = Lab Environment, D = Dim Environment, B = Bright Environment}
\end{center}
\end{table}

\subsection{Face Video Compression}
\label{compression}
The raw data of our VIPL-HR is about 1.05 TB in total, making it inconvenient for distribution. In order to facilitate access to our dataset by the other researchers, we investigate to make a compressed and resized version of our database. As studied in~\cite{mcduff2017impact}, video compression may have big influence on video-based heart rate estimation. Therefore, we carefully compare the individual video compress methods, and frame resizing methods.

We considered five different video codecs, i.e., `MJPG', `FMP4', `DIVX', `PIM1' and `X264'\footnote{\url{http://www.fourcc.org/}}, which are commonly used in the video coding community.
The resizing scales we considered are 1/2, 2/3, and 3/4 for each dimension of the original frame. All the videos are compressed using OpenCV\footnote{\url{https://opencv.org/}}  with FFmpeg interface\footnote{\url{http://ffmpeg.org/}}.
We choose a widely used remote HR estimation method Haan2013~\cite{de2013robust} as a baseline HR estimation method to verify how data compression influence its HR estimation accuracy.

\begin{figure}
\centering
\subfigure[]{
\includegraphics[width=0.42\linewidth]{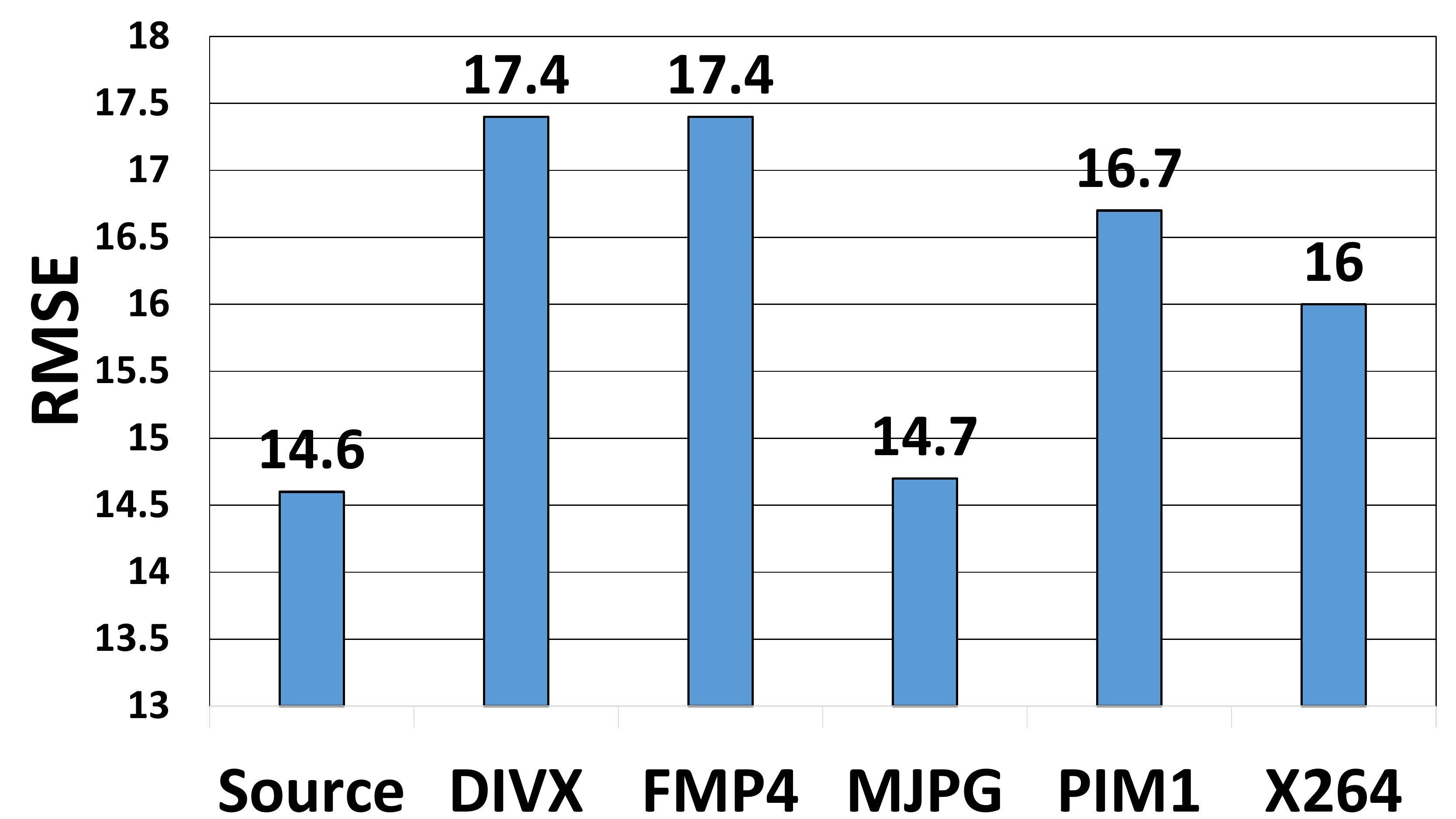}
}
\subfigure[]{
\includegraphics[width=0.42\linewidth]{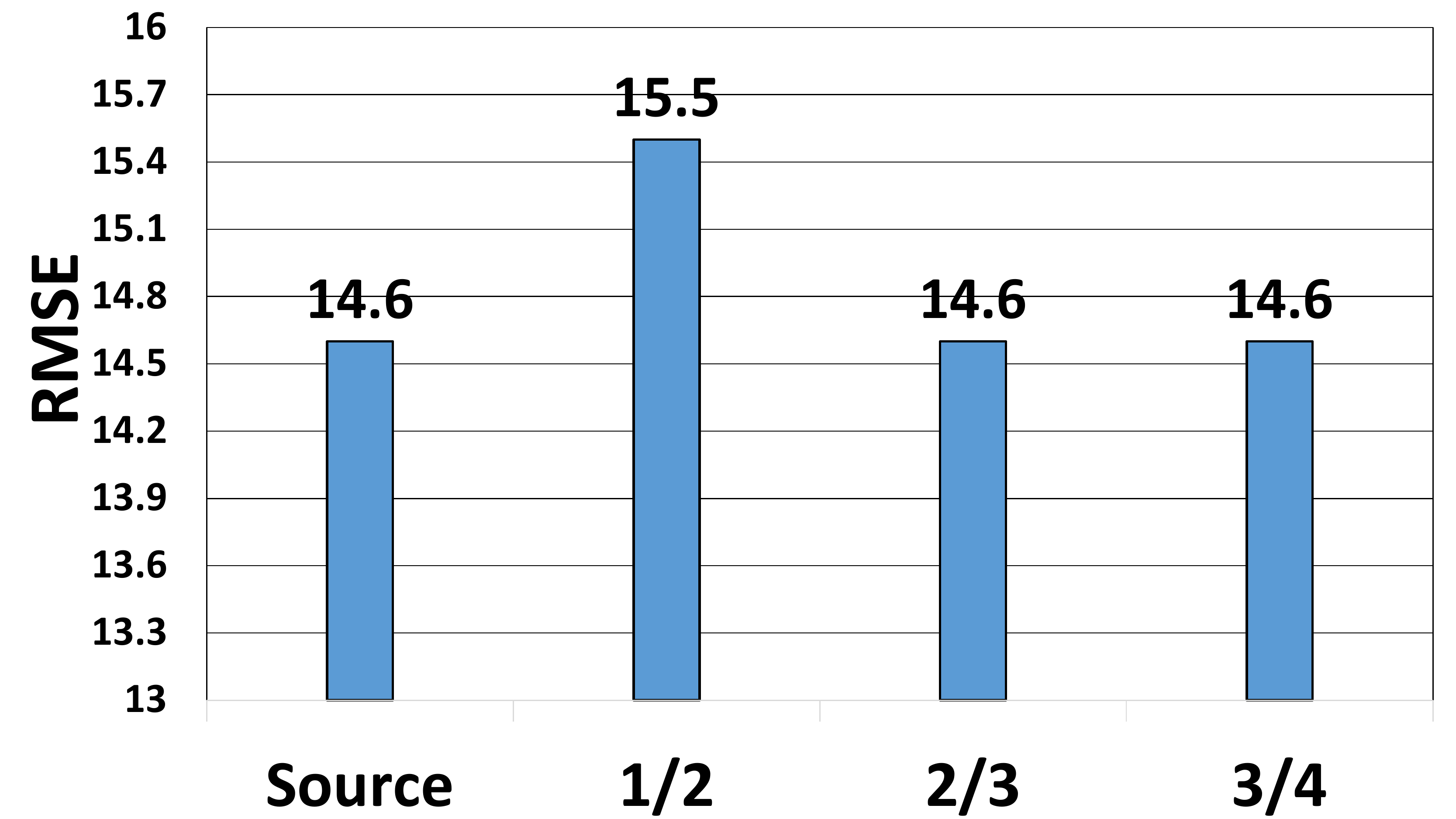}
}
\caption{Evaluations on VIPL-HR using (a) different video compression codecs and (b) different image resolutions for data compression. All tests are based on Haan2013~\cite{de2013robust}.}
\label{fig:video_compression}
\end{figure}

The HR estimation accuracies in terms of root mean square error (RMSE) by the baseline HR estimator~\cite{de2013robust} on various compressed videos are reported in Fig.~\ref{fig:video_compression}. From the results, we can see that the `MJPG' video codec works better in maintaining the HR signal in the videos while it is able to reduce the size of the database significantly. Resizing the frames to less than two-thirds of the original image resolution leads to damage to the HR signal. Therefore, we choose the `MJPG' codec and 2/3 resizing as our final data compression solution. Finally, we obtained a compressed VIPL-HR dataset with about 48 GB. We would like to share both the uncompressed and compressed databases to the research community based on individual researchers' preference.

\subsection{Database Statistics}

\begin{figure}
\centering
\subfigure[]{
\includegraphics[width=0.18\linewidth, height = 0.19\linewidth]{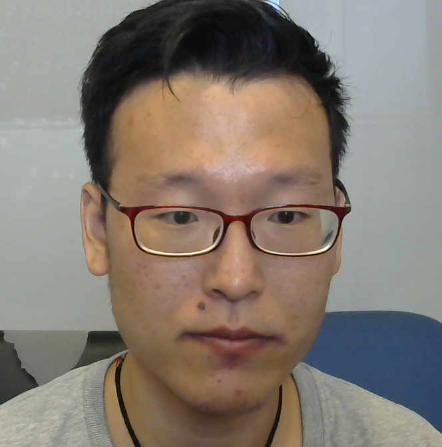}\vspace{2pt}
}
\subfigure[]{
\includegraphics[width=0.18\linewidth, height = 0.19\linewidth]{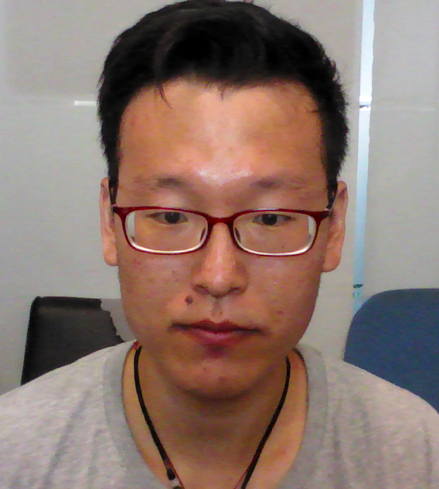}\vspace{2pt}
}
\subfigure[]{
\includegraphics[width=0.18\linewidth, height = 0.19\linewidth]{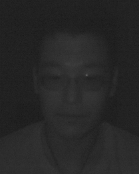}\vspace{2pt}
}
\subfigure[]{
\includegraphics[width=0.18\linewidth, height=0.19\linewidth]{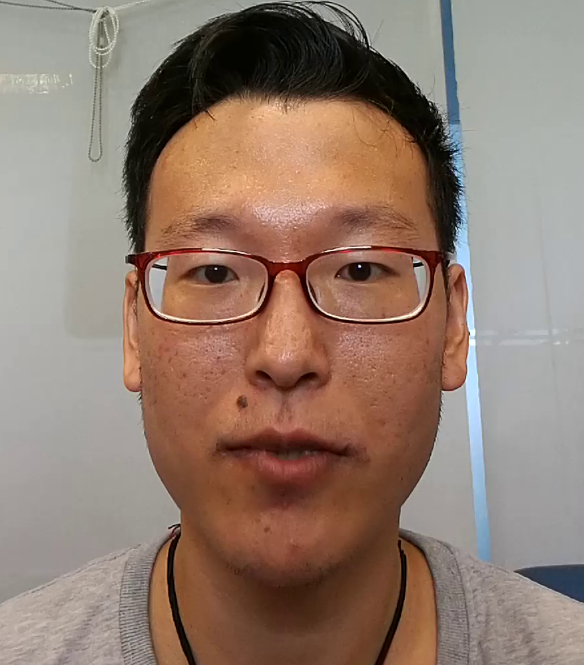}\vspace{2pt}
}
\caption{Example video frames captured by different devices: (a) Logitech C310, (b) RealSense F200 color camera, (c) RealSense F200 NIR camera, and (4) HUAWEI P9 frontal camera.}
\label{fig:camera_image}
\end{figure}

The VIPL-HR dataset contains a total of 2,378 color videos and 752 NIR videos from 107 participants with 79 males and 28 females, and ages distributed between 22 and 41. Each video is recorded with a length of about 30s, and the frame rate is about 30fps (see Table~\ref{table:database}). Some example video frames of one subject captured by different devices are shown in Fig.~\ref{fig:camera_image}.

\begin{figure}[t]
\centering
\subfigure{
\includegraphics[width=0.42\linewidth]{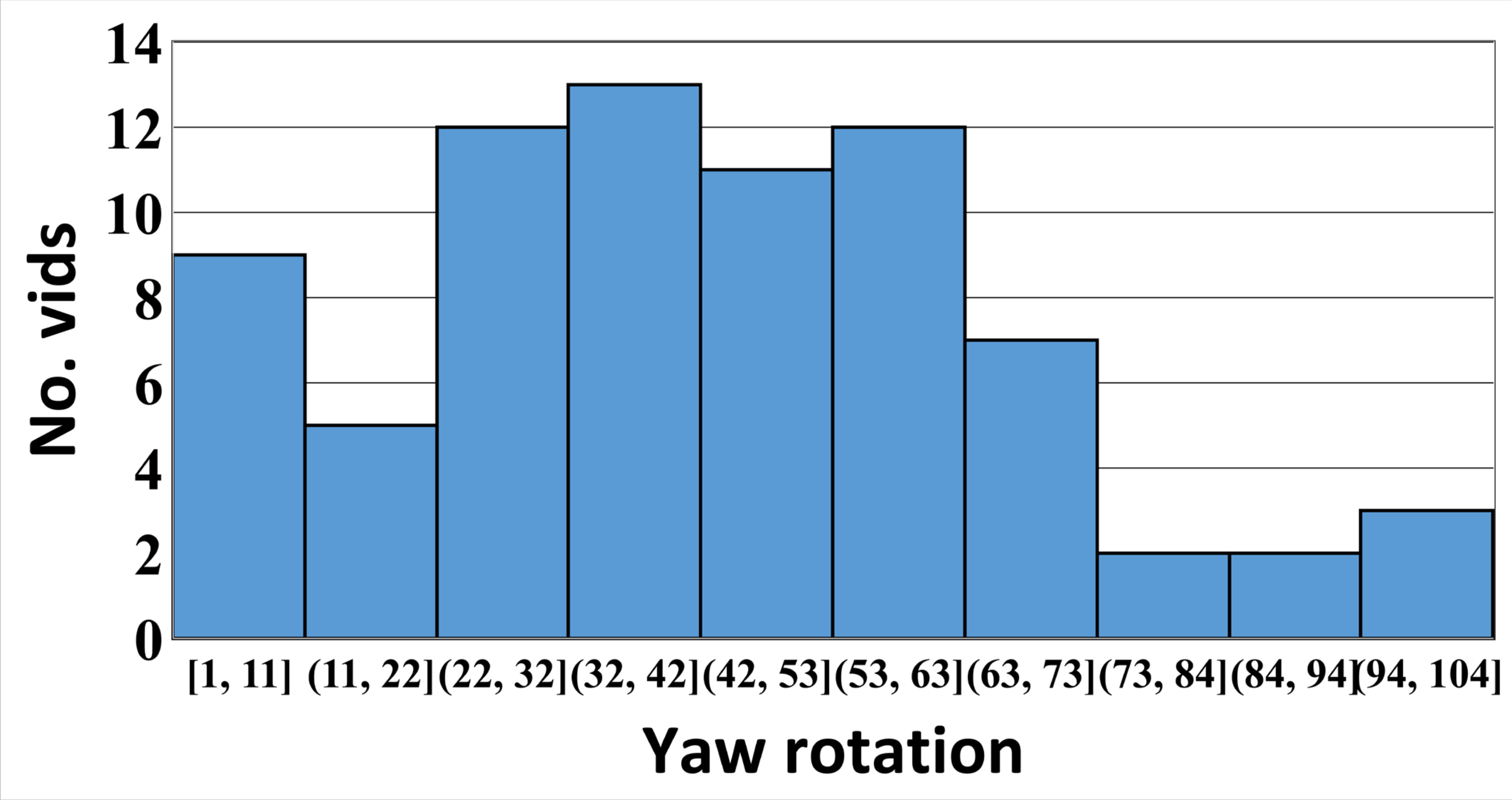}
}
\subfigure{
\includegraphics[width=0.42\linewidth]{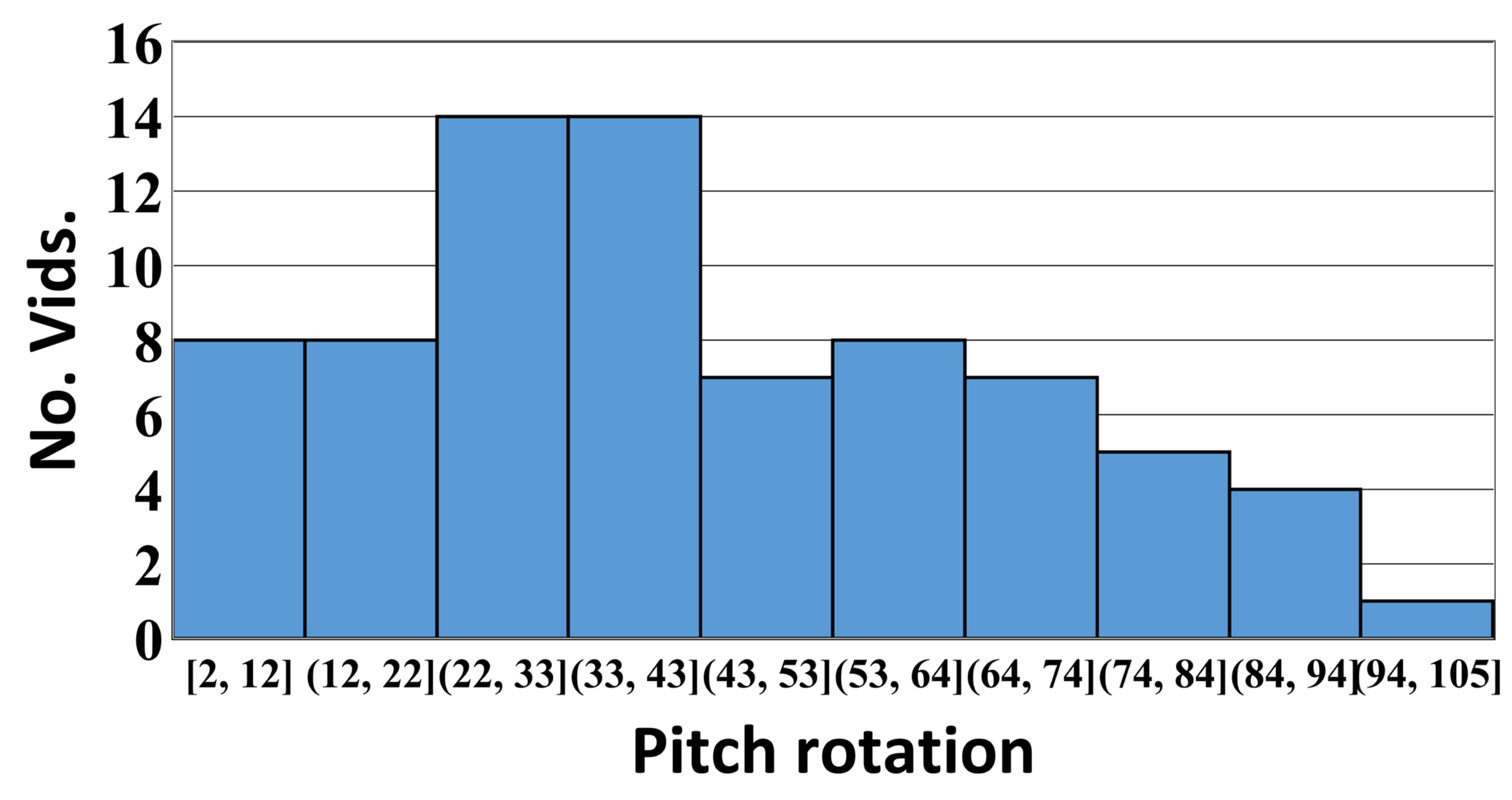}
}
\subfigure{
\includegraphics[width=0.42\linewidth]{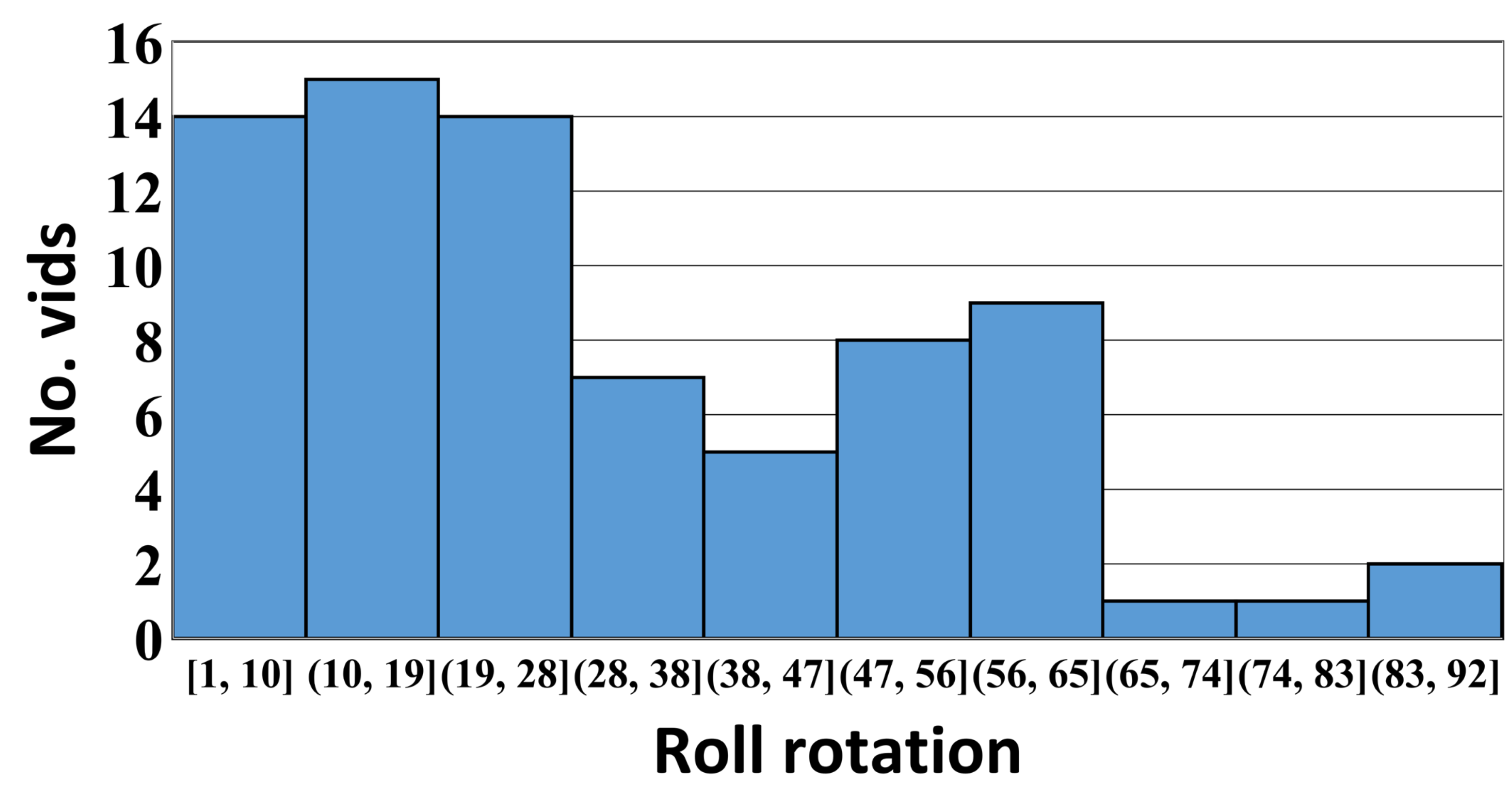}
}
\caption{ Histograms of the maximum amplitudes (in degree) of the yaw, pitch, and roll rotations for all the videos with head movement in VIPL-HR.}
\label{fig:head_motion}
\end{figure}

To obtain additional statics about the pose variations in our VIPL-HR database, we estimate head poses (yaw, pitch, and roll angles) for each video using the OpenFace head pose estimator\footnote{\url{https://github.com/TadasBaltrusaitis/OpenFace}} (see Situation 2 in Table~\ref{table:record_situation}). Histograms for maximum amplitudes of yaw, pitch, and roll rotations for all the videos can be found in Fig.~\ref{fig:head_motion}. From the histograms, we can see that the maximum rotation amplitudes of the subjects vary in a large range, i.e., the maximum rotation amplitudes are $92^\circ$ in roll, $105^\circ$ in pitch and $104^\circ$ in yaw. This is reasonable because every subject is allowed to look around during the video recording process.

\begin{figure}
\centering
\includegraphics[width=0.65\linewidth]{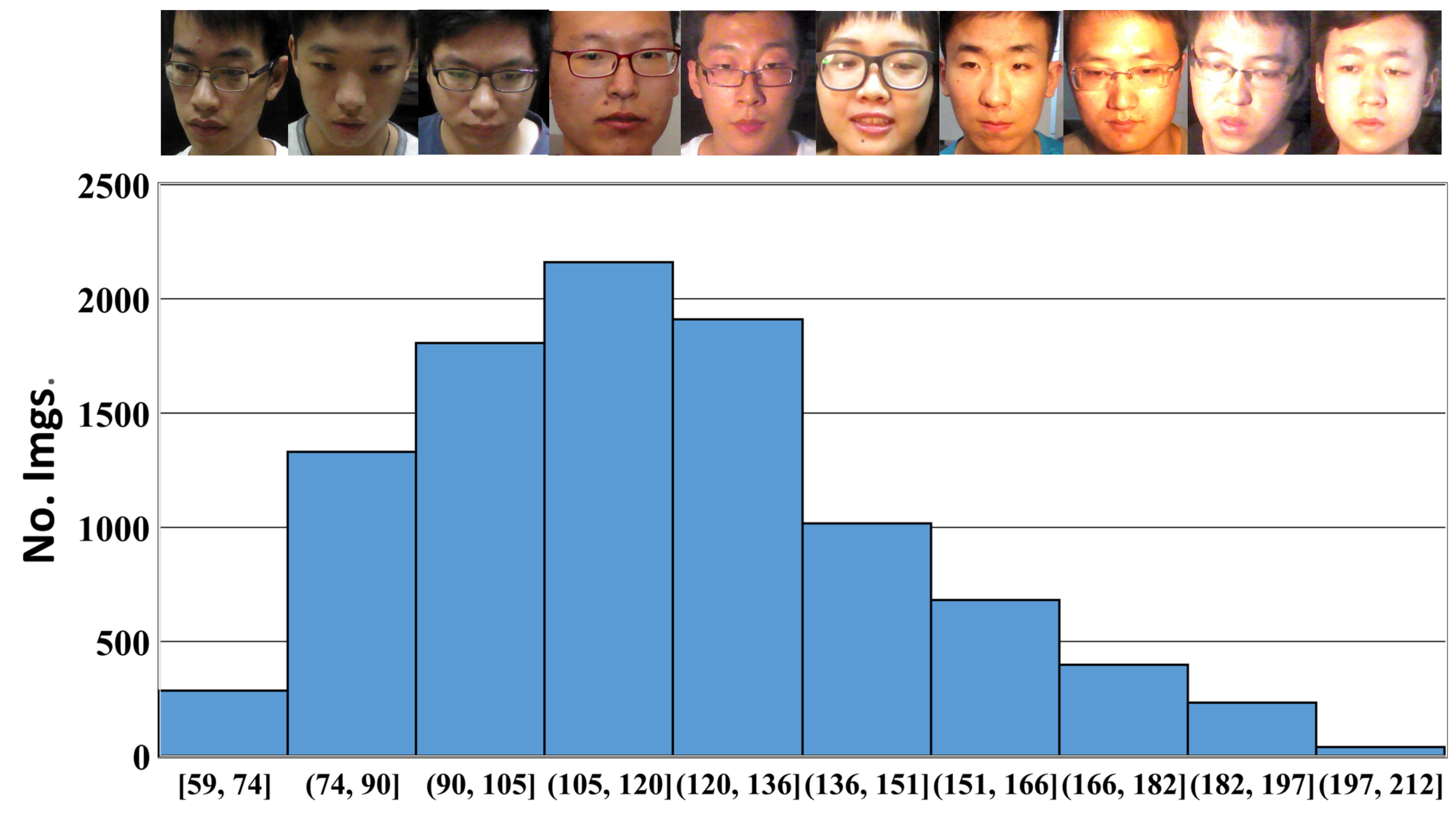}
\caption{ The histogram of the average image intensity (gray-scale) for the videos recorded under the illumination-specific situations in VIPL-HR. }
\label{fig:illumination}
\end{figure}

At the same time, in order to quantitatively demonstrate the illumination changes in VIPL-HR database, we have calculated the mean gray-scale intensity of face area for Situation 1, Situation 4, and Situation 5 in Table~\ref{table:record_situation}. The results are shown in Fig.~\ref{fig:illumination}. We can see that the mean gray-scale intensity varies from 60 to 212, covering complicated illumination variations.

\begin{figure}[t]
\centering
\includegraphics[width=0.65\linewidth]{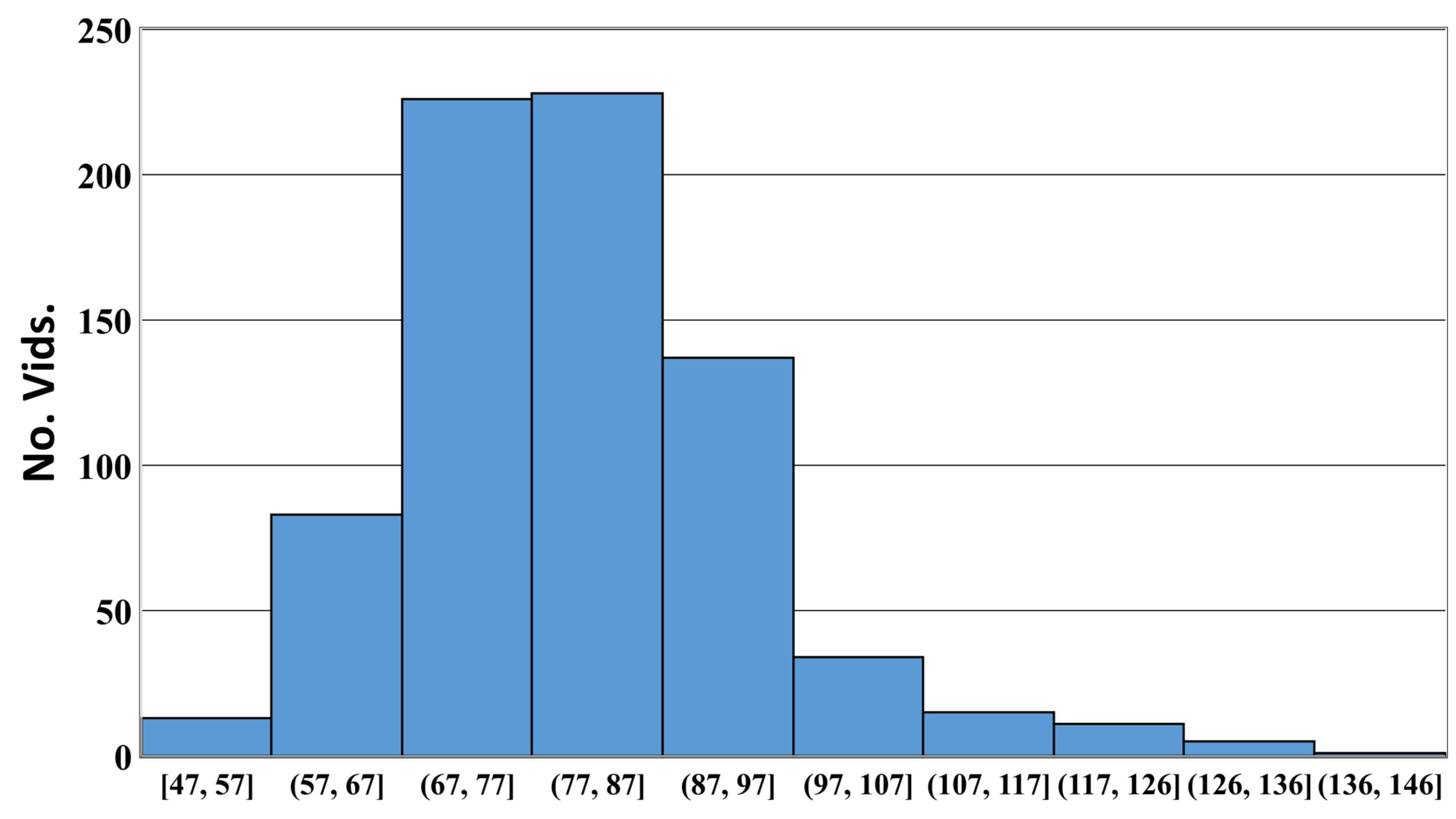}
\caption{ The histogram of the ground-truth HR distribution in VIPL-HR. }
\label{fig:gt_HR}
\end{figure}

A histogram of ground-truth HRs is also shown in Fig.~\ref{fig:gt_HR}. We can see that the ground-truth HRs in VIPL-HR vary from 47 bpm to 146bpm, which covers the typical HR range\footnote{\url{https://en.wikipedia.org/wiki/Heart_rate}}. The wide HR distribution in VIPL-HR mitigates the gap between the HR range covered in dataset collection and the HR distribution that can appear in daily-life scenes. The relatively large size of our VIPL-HR also makes it possible to leverage deep learning methods to build more robust data-driven HR estimation model.

\section{Proposed Approach}
\label{Proposed_Approach}

\begin{figure*}[t]
      \centering
      \includegraphics[width=0.65\linewidth]{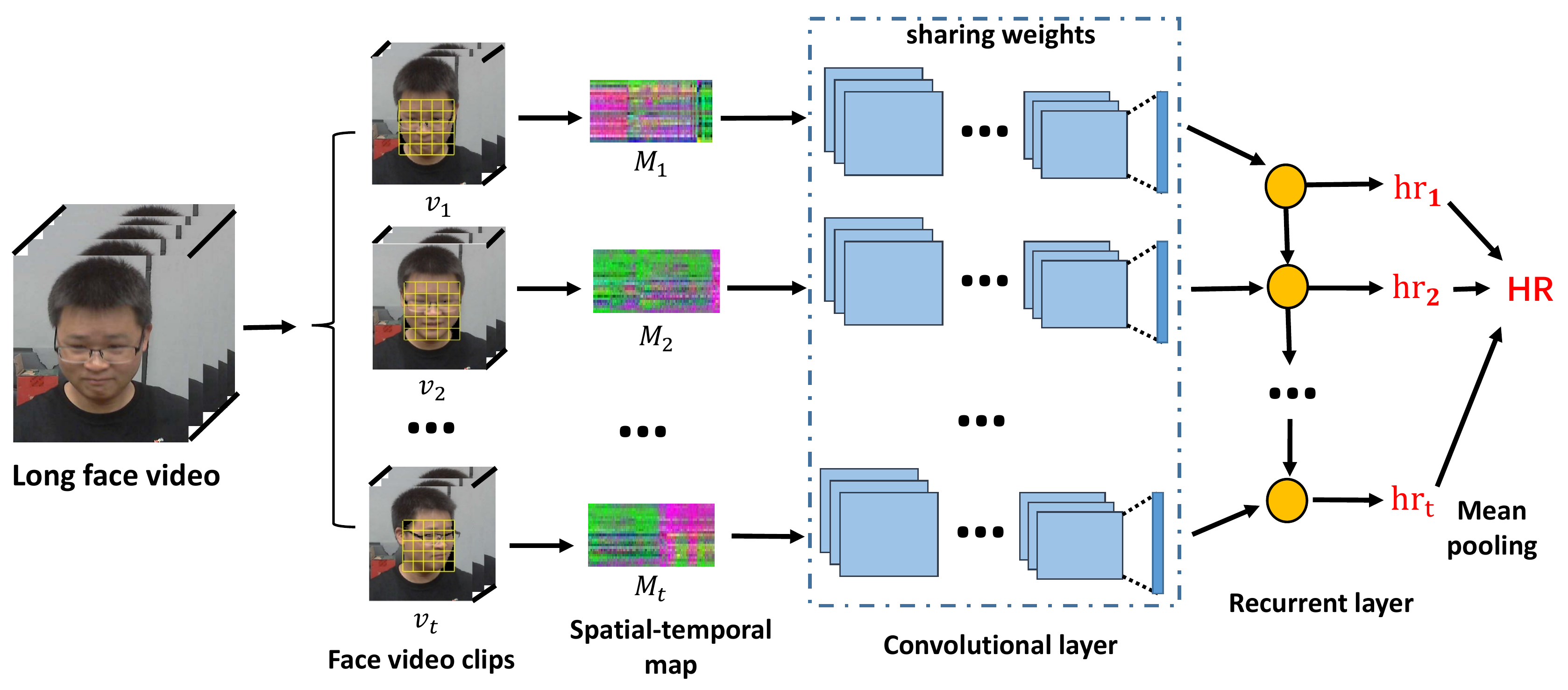}
      \caption{Overview of our RhythmNet. Given an input video sequence, we first divide it into multiple short video clips ($v_1, v_2, \cdots, v_t$), and perform face and landmark detection on each frame for face alignment. Then spatial-temporal maps are generated from the aligned face images per video clip to represent the HR signals, and a deep network consisting of convolutional layers and recurrent layers is trained to predict the HR from the spatial-temporal maps. Finally, the HR estimated for the input video sequence is computed as the average of all the estimated HRs from individual video clips. }
      \label{fig:overview}
\end{figure*}

Prior work performed remote HR estimation mainly based on hand-crafted features, which are built based on certain assumptions of the HR signal and environment. These methods may fail in realistic situations when the assumptions do not hold. Given the large-scale VIPL-HR database with diversities in illumination, pose, and sensors, we are able to build a data-driven HR estimation model using deep learning methods. In this section, we are going to introduce the RhythmNet for remote HR estimation from face video. Fig.~\ref{fig:overview} gives an overview of our RhythmNet. We first perform detection and facial landmark detection to locate region of interest (ROI) facial areas, from which a spatial-temporal map is computed as a novel low-level representation of the HR signal. Finally, a CNN-RNN model is designed to learn a mapping from the spatial-temporal maps to the HR.

\subsection{Face Detection, Landmark Detection and Segmentation}
\label{preprocessing}

We use an open source face detector SeetaFace\footnote{\url{https://github.com/seetaface/SeetaFaceEngine}} to detect the face and localize 81 facial landmarks (see Fig.~\ref{fig:SignalMap}). Since face and facial landmarks detection is able to run at a frame rate of more than 30 fps, we perform face detection and landmarks detection on every frame in order to get accurate and consistent ROI facial areas in a video sequence. A moving average filter is also applied to the 81 facial landmarks across frames to get more stable landmark locations.

In order to make full use of all the informative parts containing the color changes due to heart rhythms, we choose to use the whole face area for further processing. Face alignment is firstly performed using the eye centre points, and then a face bounding box is defined with a width of $w$ (where $w$ is the horizontal distance between the outer cheek border points) and height $1.2*h$ (where $h$ is the vertical distance between chin location and eye eyebrow centre). Skin segmentation is then applied to the defined ROI to remove the non-face area such as eye region and background area. Since resizing may introduce noise to the HR signals, we choose to use the original face region cropped based on facial landmarks for further processing.

\subsection{Spatial-temporal Map for Representing HR Signals}
\label{Signalmap}

\begin{figure*}
      \centering
      \includegraphics[width=0.7\linewidth]{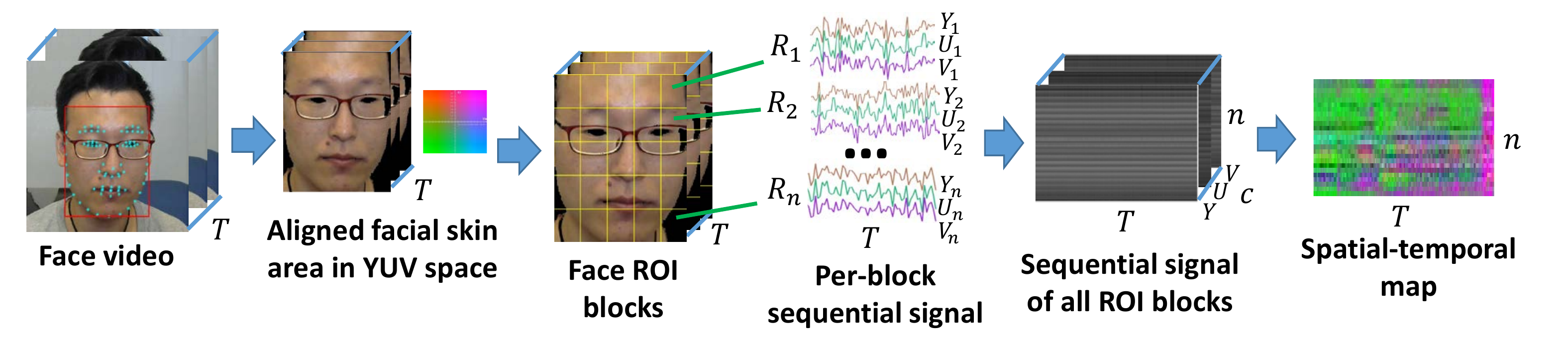}
      \caption{An illustration of spatial-temporal map generation from face video. We first align faces across different frames based on the detected facial landmarks, and transform the aligned face images to YUV color space. Then the facial area is divided into $n$ ROI blocks $R_1, R_2,\cdots,R_n$, and the average color value is computed for each color channel within each block. The per channel average color values at the same block location but different frames                                                                                                                                                                                                                                                                                                                                                                                                                                                                                                                                                                                                                                                                                                                                                                                                                                                                                                                                                                                                                                                                                                                                                                                                                                                                                                                                                                                                                           are concatenated into a sequences, i.e., $Y_1, U_1, V_1, Y_2, U_2, V_2,\cdots, Y_n, U_n, V_n$. These $n*c$ sequences are placed into rows to form a spatial-temporal map with the size of $T \times n\times c$.}
      \label{fig:SignalMap}
\end{figure*}

The \textbf{only} useful information for rPPG-based HR estimation in face video is the skin color variations caused by variations in volume and oxygen saturation of the blood in the vessels due to heart beats, which are very minor in amplitude and can be affected by head movements, illumination variations and sensor noises. In order to suppress noises and improve the SNR of HR signals, Most rPPG-based HR estimation methods use the average pixel values of RGB channels of the whole face as the HR signal representation~\cite{poh2010non,poh2011advancements,de2013robust,wang2017algorithmic,feng2015motion}. Such an average pooling representation provides better robustness than each single pixel.
This can be regarded as the empirical rule, and can work smoothly for rPPG-based HR estimation under various conditions.
Unlike the existing methods which only use the average pooling of ROI blocks to compute the HR signal, we have proposed a spatial-temporal map (see Fig.~\ref{fig:SignalMap}) to highlight the heart rhythm signals while suppressing the other information irrelevant to heart rhythm. Such a spatial-temporal map can enforce the input to the succeeding network to be as specific as to the heart rhythm signal, and make it possible for us to leverage CNN to learn informative representation for the final HR estimation task.

Specifically, for a video clip with $T$ frames and $c$ color space dimensions, we first get the face areas for all the frames as stated in~\ref{preprocessing}. Then, the face area in each frame is divided into $n$ ROI blocks $R_1, R_2,\cdots,R_n$.
The amplitude of these skin color variations are very weak, and average pooling has been proven to be effective in removing noises in rPPG-based HR estimation methods.
Let $C(x, y, t)$ denote the value at location $(x, y)$ of the $t^{th}$ frame from different dimensions of the color space, and the average pooling of the $i^{th}$ ROI block of the $t^{th}$ frame for each channel can be stated as
\begin{equation}
\overline{C}_{i}(t) = \frac{\sum_{x,y\in ROI_{i}}C(x,y,t)}{|ROI_{i}|}
\end{equation}
where $|ROI_{i}|$ denotes the area of a block ROI (the number of pixels). So, for each video clip we can obtain $3\times n$ temporal sequences with the length of $T$ for different dimensions of the color space, e.g., $\mathbf{C}_{i} = \{\overline{C}_{i}(1),\overline{C}_{i}(2),\cdots,\overline{C}_{i}(T)\}$, where $C$ donates one of the $c$ color space dimensions and $i$ donates the index of the ROI. In order to make the best use of the HR signals, a min-max normalization is applied to each temporal signal, and the values of the temporal series are scaled into [0, 255]. Then, we place the $n$ temporal sequences into rows and obtain a spatial-temporal map representation for the HR signal in a video clip. Eventually, we get a spatial-temporal representation from the raw face video clip with the size of $T\times n\times  c$ as the input of our succeeding deep HR estimation network.

Color space selection is very important for representing the HR signal. As stated in~\cite{tsouri2015benefits}, alternative color spaces derived from RGB video are beneficial for getting a better HR signal representation. Therefore, we take two kinds of color spaces into consideration: i) color spaces with dimension directly related to color, such as HSV color space, and ii) color spaces which are commonly used for skin segmentation, i.e., YUV and YCrCb color spaces. Testing on different color spaces can be found in Section~\ref{Key_components_analysis}. We imperially choose to use the YUV color space (see our evaluations in Section~\ref{Key_components_analysis}). The color space transformation can be formulized as
{\fontsize{8pt}\baselineskip\selectfont\begin{align}
\left[
\begin{array}{c}
Y \\
U \\
V \\
\end{array}
\right]
= \left[
\begin{array}{ccc}
0.299 & 0.587 & 0.114\\
-0.169 & -0.331 & 0.5\\
0.5 & -0.419 & -0.081\\
\end{array}
\right]
\left[
\begin{array}{c}
R \\
G \\
B \\
\end{array}
\right] + \left[
\begin{array}{c}
0 \\
128 \\
128 \\
\end{array}
\right]
\end{align}}

Another situation we need to consider is face detection failures for a few frames, which may happen when the subject's head moves or rotates too fast. This will lead to missing data of HR signals and the spatial-temporal representation. In order to handle this issue, we randomly mask a small part of the spatial-temporal maps along the time dimension to simulate the missing data cases, and use the partially masked spatial-temporal maps as augmented data to enhance the robustness of our RhythmNet.

\subsection{Temporal Modeling for HR measurement}

For each face video, we divide it into individual video clips, i.e., $v_1, v_2, v_t$, using a fixed sliding window containing $w$ frames moved with a step of 0.5 seconds. The video clips are then used for computing the spatial-temporal maps of each clip, which are used as the input to our succeeding network.
We choose ResNet-18~\cite{he2016deep} as the backbone convolutional layers, which includes four blocks made up of convolutional layers and residual link, one convolutional layer, and one fully connected layer for the final regression. $L_1$ loss is used for measuring the difference between the predicted HR and ground truth HR. The output of the network is a single HR value regressed by a fully connected layer. All the estimated HRs from individual video clips are normalized based on the frame rate of the face video.

We further take the relationship between adjacent measurements of two video clips into consideration and utilize a Gated Recurrent Unit (GRU)~\cite{cho2014learning,niu2018automatic} consisting of a cell, a reset gate and an update gate to model the temporal relationship between succeeding measurement. To be specific, the features extracted from the backbone CNN are fed to a one-layer GRU structure. The output of GRU is fed into a fully connected layer to regress the HR values for individual video clips. For each face video, the average of all the predicted HRs for individual video clips are computed as the final HR result.

Since the variance of the subjects' HRs is small during a very small period of time~\cite{niucontinuous}, we introduce a smooth loss function to constrain the smoothness of adjacent HR measurements. We take $T$ continuous measurements $hr_{1}, hr_2, \cdots, hr_{T}$ into consideration, which are estimated from the GRU for a small period of time, i.e., 3 seconds. The mean HR within the $T$ continuous measurements is then computed as
$ hr_{mean} = \frac{1}{T} \sum_{i=1}^{T} hr_{t} $ and the smooth loss $L_{smooth}$ is defined as
\begin{equation}
L_{smooth} = \frac{1}{T} \sum_{i=1}^{T} \|hr_{t} - hr_{mean}\|
\end{equation}
During the backpropagation, the partial derivative of the smooth loss $L_{smooth}$ with respect to the input $hr_{t}$ can be
computed as
\begin{equation}
\begin{split}
\frac{\partial L_{smooth}}{\partial hr_{t}} =& (\frac{1}{T}-1)sgn(hr_{mean} - hr_{t}) + \\
& \sum_{i=1, i\neq t}^{T} \frac{1}{T} sgn(hr_{mean} - hr_{t})
\end{split}
\end{equation}
where $sgn(x)$ is a sign function. The final loss function can be written as
\begin{equation}
\label{equ:lambda}
L = L_{l1} + \lambda L_{smooth}
\end{equation}
where $L_{l1}$ denotes the $L_1$ loss function and $\lambda$ is a parameter for balancing the two terms.

While RNN is widely used for modeling the temporal continuity in object tracking, action recognition, etc., its effectiveness in modeling the temporal relationships of physiological signals (such as HR), is not known. Therefore, our work is the first known approach uses RNN to improve the temporal estimation stability of HR.

\section{Experiments}
\label{experiments}

In this section, we provide evaluations of the proposed RhythmNet covering three aspects: i) intra-database testing, ii) cross-database testing, and iii) key components analysis.

\subsection{Database and Experimental Setting}
\label{protocol}

We provide evaluations on two widely used public-domain databases, i.e., MAHNOB-HCI~\cite{soleymani2012multimodal} and MMSE-HR~\cite{Tulyakov2016Self} as well as the VIPL-HR database we collected. The details about these databases can be found in Table~\ref{table:database}. The ground-truth HRs of MAHNOB-HCI and MMSE-HR database are computed from the electrocardiography (ECG) signals provided in the databases using the OSET ECG Toolbox\footnote{\url{http://www.oset.ir}}.

Different metrics have been used in the literature for evaluating the HR estimation performance, such as the mean and standard deviation (Mean and Std) of the HR error, the mean absolute HR error (MAE), the root mean squared HR error (RMSE), the mean of error rate percentage (MER), and Pearson's correlation coefficients $r$~\cite{li2014remote,Tulyakov2016Self}. We also use these evaluation metrics in our experiments below.

For the MMSE-HR and VIPL-HR datasets, we use a temporal sliding window with $300$ frames to compute the spatial-temporal maps. For the MAHNOB-HCI database, since its original frame rate is 61 fps, we first downsample the videos to 30.5 fps before using the same temporal window size. The number of the ROI blocks used for spatial-temporal map generation for all the database is 25 ($5\times 5$ grids). During the training phase, half of the generated spatial-temporal maps were randomly masked, and the mask length varies from $10$ frames to $30$ frames. For the temporal relationship modelling, six adjacent estimated HRs are used to compute the $L_{smooth}$.
The balance parameter $\lambda$ in Equ.~\ref{equ:lambda} is set to 100. Since the face sizes in the NIR videos are small for the face detector, only 497 NIR videos with detected faces are involved in the experiments. Our RhythmNet is implemented using PyTorch\footnote{\url{https://pytorch.org/}}. We use an Adam solver~\cite{kingma2014adam} with an initial learning rate of 0.001, and set the maximum epoch number to 50.

\subsection{Intra-database Testing}
\label{intra-database}

We first conduct intra-database experiments on each of the three databases. For the intra-database testing, in order to prove the effectiveness of the proposed HR estimator, we use a five-fold \textbf{subject-exclusive} cross-validation for the VIPL-HR database, which means that the subjects in the training set will not appear in the testing set. Since the MAHNOB-HCI and MMSE-HR databases have very limited face videos, we use a three-fold subject-independent cross-validation protocol.

\subsubsection{Experiments on Color Face Videos}
We first perform intra-database evaluations using RGB face videos in our large VIPL-HR database, and compare the proposed approach with a number of state-of-the-art traditional methods (Haan2013~\cite{de2013robust}, Tulyakov2016~\cite{Tulyakov2016Self}, POS~\cite{wang2017algorithmic}). Two deep learning based methods, i.e., I3D~\cite{carreira2017quo} and DeepPhy~\cite{chen2018deepphys}, are also used for comparison. I3D [37] is widely used for video analysis, and DeepPhy [27] is the state-of-the-art deep learning based HR estimation method. For I3D, we directly use the code provided by DeepMind\footnote{\url{https://github.com/deepmind/kinetics-i3d}}, and trained it for HR estimation task. For DeepPhy, we contacted the authors, but the code is not publicly available; so we have re-implemented DeepPhy based on the descriptions in [27]. We use the same five-fold subject-exclusive protocol for all the deep learning based methods. For our RhythmNet, we report both results using and without using GRU for comparisons with the traditional state-of-the-art methods.
The results of individual HR estimation methods are given in Table~\ref{table:vipl_color}.

\setlength{\tabcolsep}{2pt}
\begin{table}
\begin{center}
\caption{The HR estimation results by the proposed approach and several state-of-the-art methods on color face videos of the VIPL-HR database.}
\label{table:vipl_color}
\begin{tabular}{lcccccc}
\toprule
\multirow{2}{*}{Method} & Mean  & Std  & MAE & RMSE  & \multirow{ 2}{*}{MER} & \multirow{ 2}{*}{$r$}\\
&(bpm) &(bpm) &(bpm)&(bpm)\\
\toprule
Tulyakov2016~\cite{Tulyakov2016Self} & 10.8 & 18.0 & 15.9 & 21.0 & 26.7\% & 0.11\\
POS~\cite{wang2017algorithmic} & 7.87 & 15.3 & 11.5 & 17.2 & 18.5\% & 0.30\\
Haan2013~\cite{de2013robust} & 7.63 & 15.1 & 11.4 & 16.9 & 17.8\% & 0.28 \\
I3D~\cite{carreira2017quo} & 1.37 & 15.9 & 12.0 & 15.9 & 15.6\% & 0.07\\
DeepPhy~\cite{chen2018deepphys} & -2.60 & 13.6 & 11.0 & 13.8 & 13.6\% &  0.11\\
\hline
\noalign{\smallskip}
RhythmNet w/o GRU & 1.02  & 8.88 & 5.79 & 8.94   & 7.38\%  & 0.73\\
RhythmNet & \textbf{0.73}  & \textbf{8.11} & \textbf{5.30} & \textbf{8.14}   & \textbf{6.71\%}  & \textbf{0.76}\\
\bottomrule
\end{tabular}
\end{center}
\end{table}

\begin{figure}
\centering
\subfigure[]{
\includegraphics[width=0.46\linewidth]{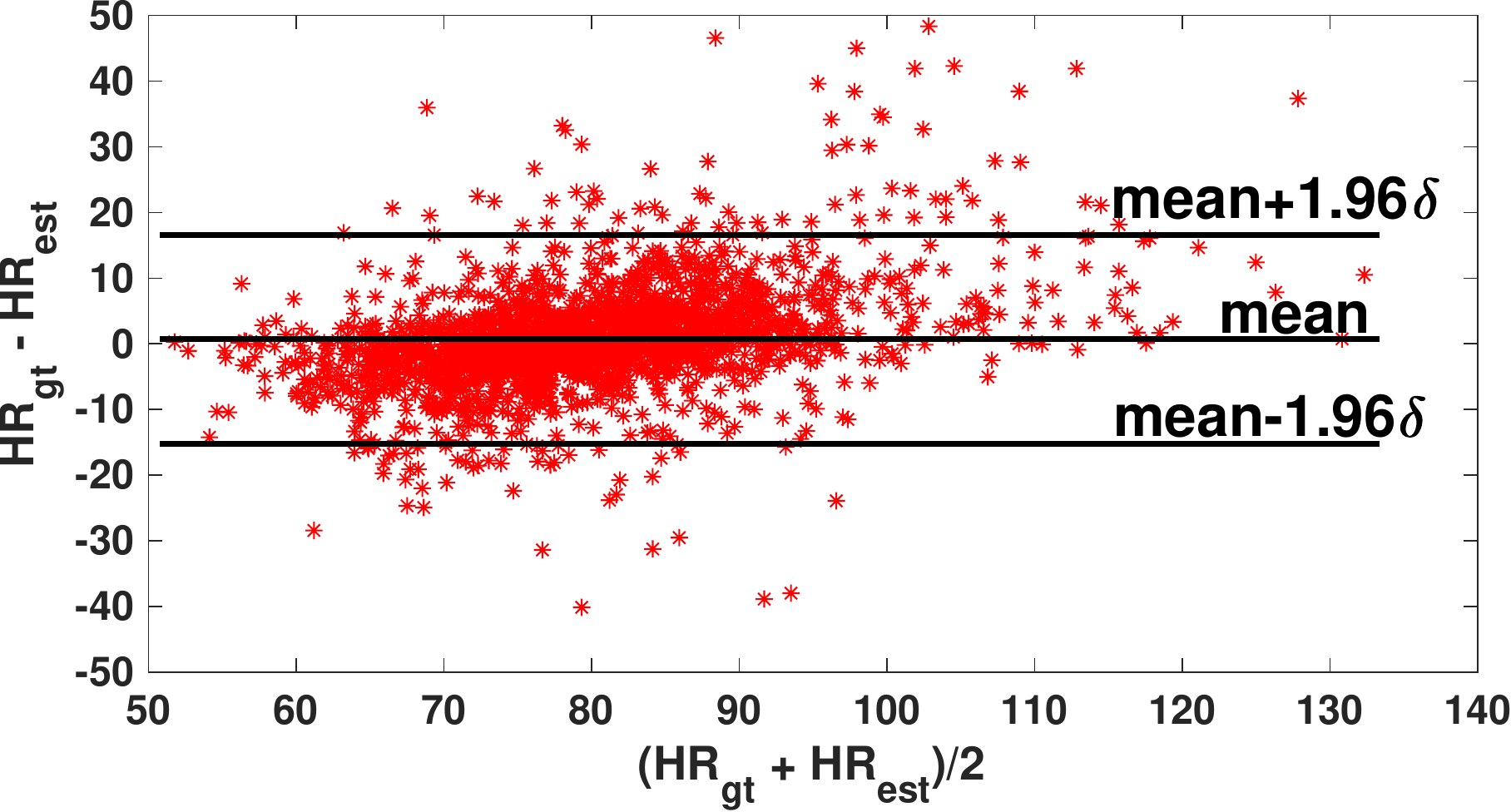}
\label{fig:BAplot_color_VIPL}
}
\subfigure[]{
\includegraphics[width=0.46\linewidth]{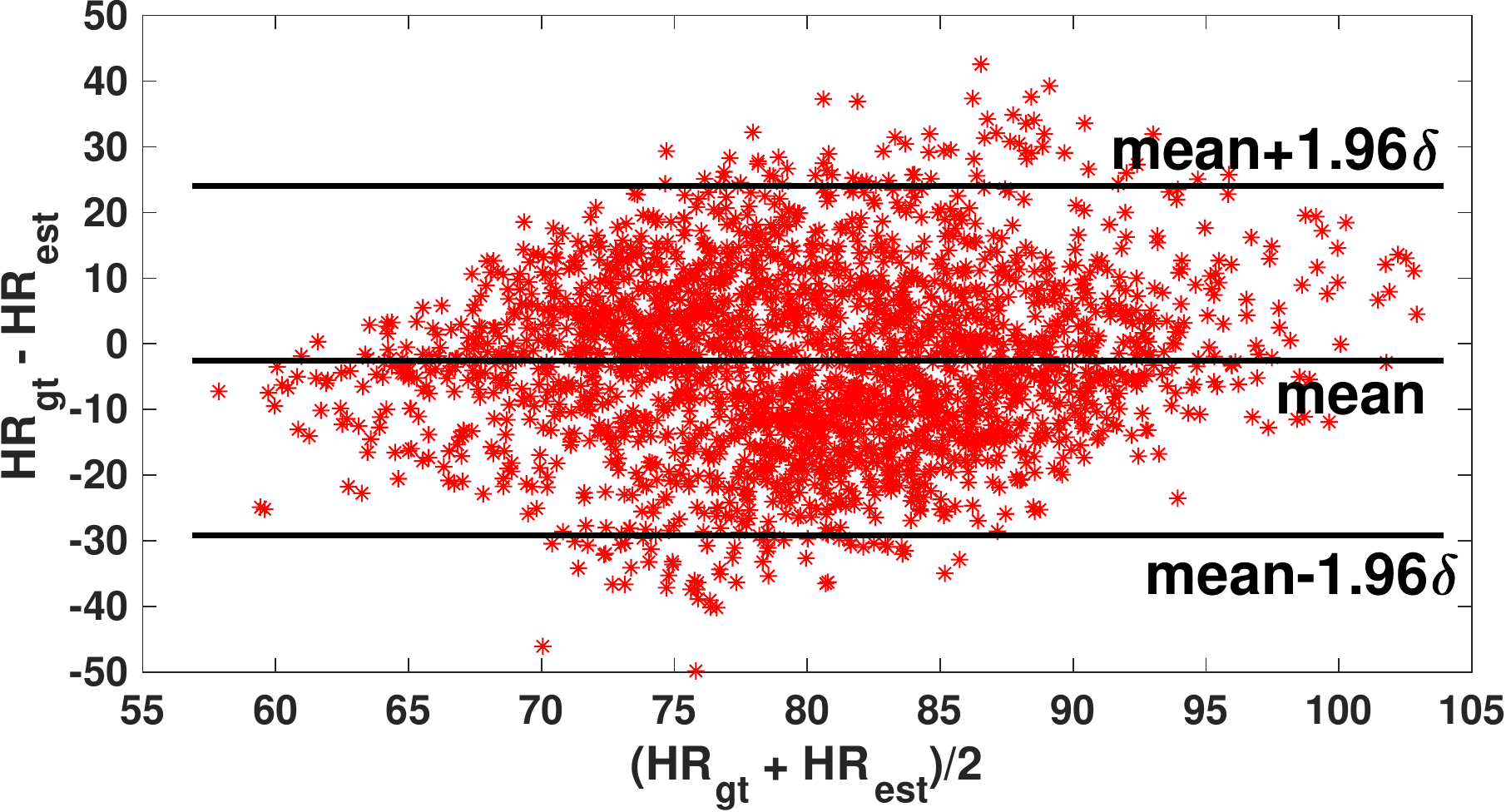}
\label{fig:BAplot_color_DeepPhy}
}
\caption{ The Bland-Altman plots demonstrating the agreement of the $HR_{est}$ and $HR_{gt}$ for (a) RhythmNet and (b) DeepPhy~\cite{chen2018deepphys} in the VIPL-HR database. The lines represent the mean and $95\%$ limits of agreement.}
\label{fig:BAplot}
\end{figure}

From Table~\ref{table:vipl_color}, we can see that the proposed approach achieves a promising result with a RMSE of 8.94 bpm when only using the CNN part of RhythmNet, which is a much lower error than the best of the
baseline methods~\cite{chen2018deepphys} (RMSE = 13.8 bpm).
The HR estimation error by the complete RhythmNet is further reduced to 8.14 bpm. In Fig.~\ref{fig:BAplot_color_VIPL}, we also draw a Bland-Altman plot of the ground-truth HR and the estimated HR by our RhythmNet to analyze the estimation consistencies by individual approaches. The Bland-Altman plot for DeepPhy~\cite{chen2018deepphys} is also given for comparison in Fig.~\ref{fig:BAplot_color_DeepPhy}. Again, it can be seen that our method achieves a better consistency than DeepPhy~\cite{chen2018deepphys} on the VIPL-HR database.

\begin{figure}
      \centering
      \includegraphics[width=0.65\linewidth]{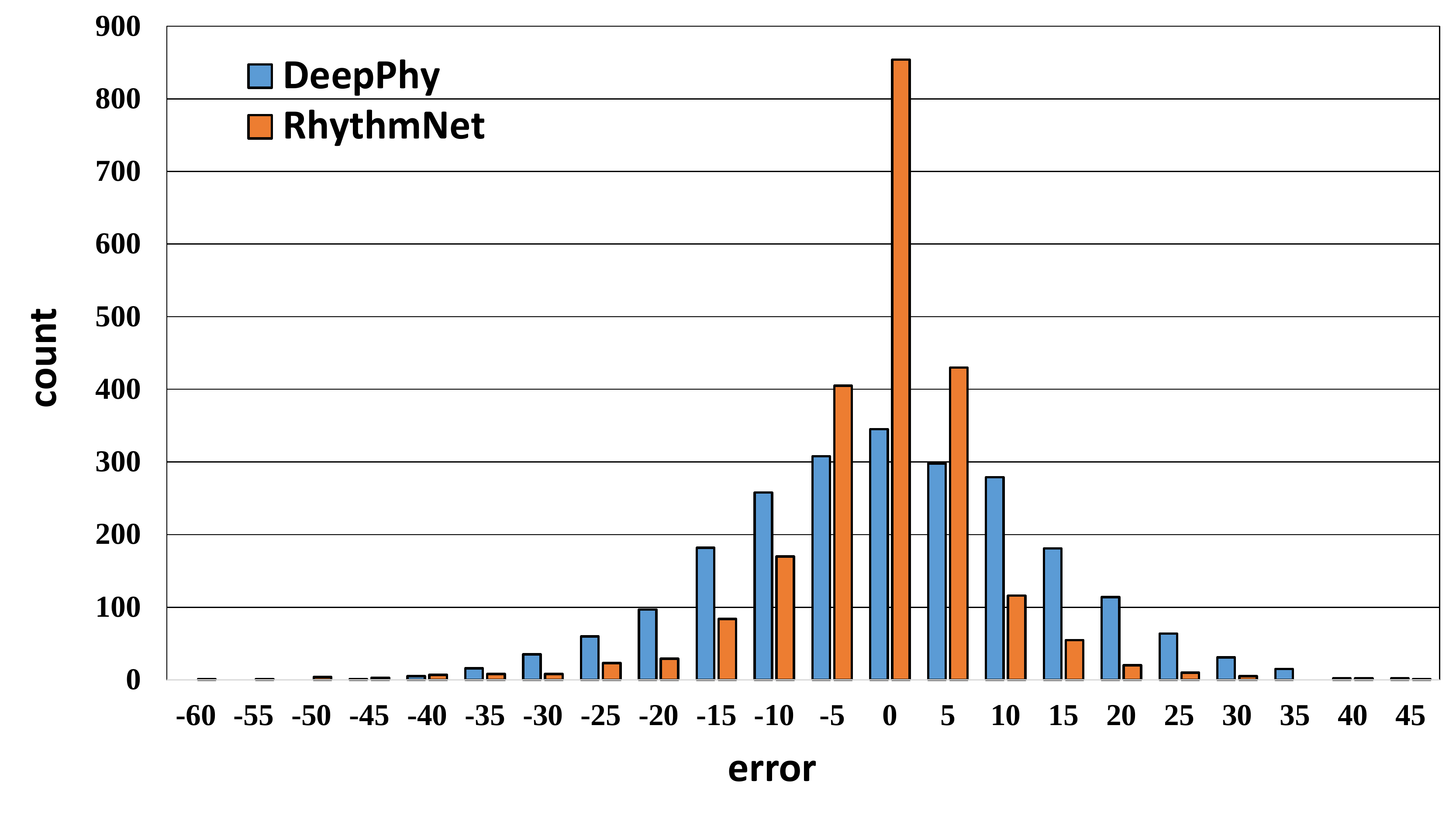}
      \caption{Comparison of the HR estimation error distributions between the proposed RhythmNet and a learning-based state-of-the-art method DeepPhy~\cite{chen2018deepphys}.}
      \label{fig:error_distribution}
\end{figure}

\begin{figure}[t]
      \centering
      \includegraphics[width=0.55\linewidth, height = 0.50\linewidth]{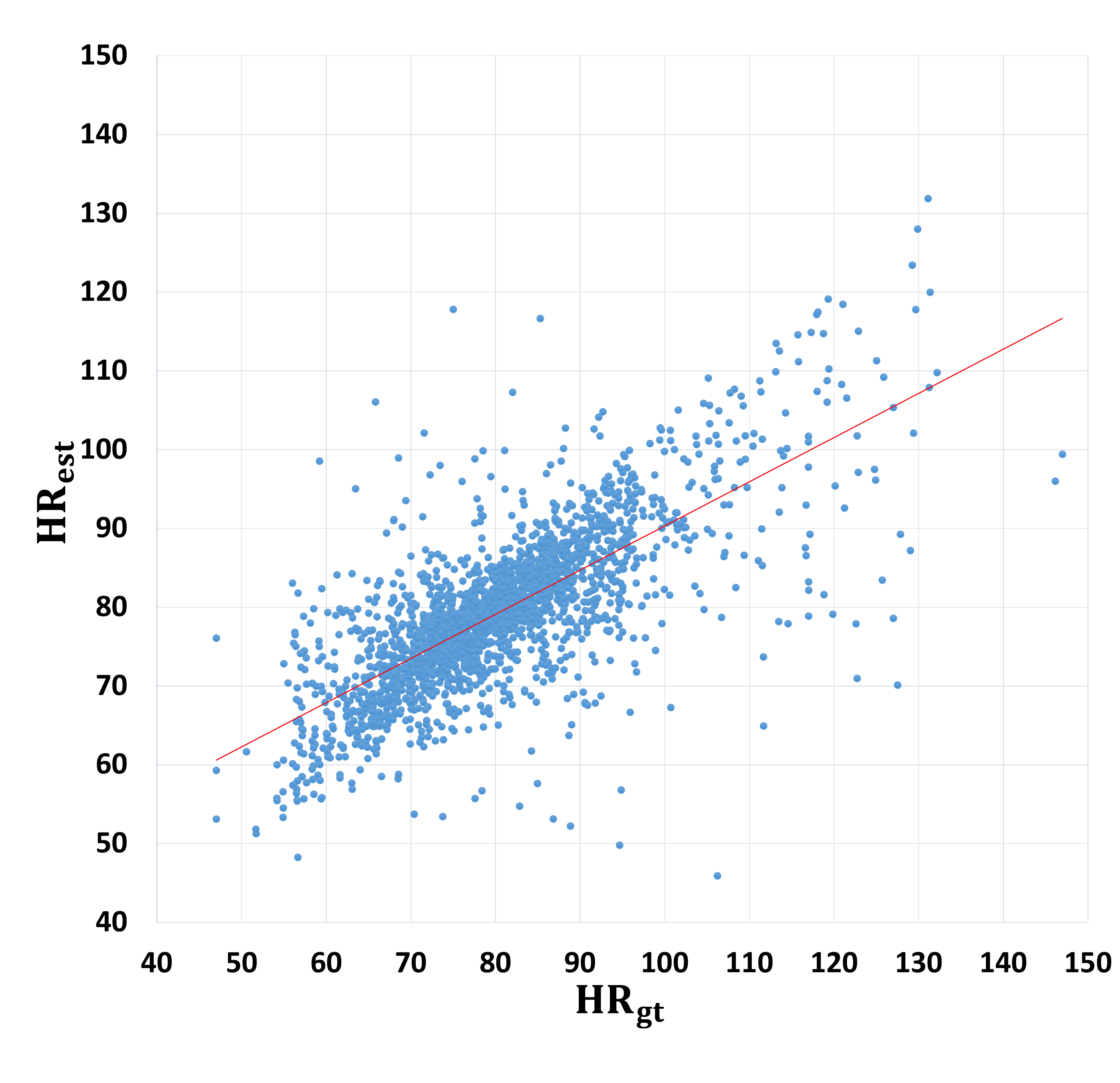}
      \caption{The scatter plot comparing the ground truth $HR_{gt}$ and the estimated $HR_{est}$ by our RhythmNet for RGB face videos on the VIPL-HR dataset.} 
      \label{fig:HR_gt_plot}
\end{figure}

We further check the HR estimation error distributions of the proposed method and~\cite{chen2018deepphys}. As shown in Fig.~\ref{fig:error_distribution}, for most of the samples (71\%), the proposed approach achieves a lower HR estimation error than 5bpm, while the percentage for~\cite{chen2018deepphys} is only 41.5\%.
We also plot the estimated HR comparing against the ground-truth HR in Fig.~\ref{fig:HR_gt_plot} to see the correlations between the estimated HR and the ground-truth HR. We can see that overall the predicted HRs are well correlated with the ground truth in a wide range of HR from 47 bpm to 147 bpm.
The results show the effectiveness of the proposed end-to-end learning-based HR estimation method in handling challenges due to pose, illumination, and sensor variations.

We can also notice from Table~\ref{table:vipl_color} that deep learning based methods (our RhythmNet, I3D~\cite{carreira2017quo}, and DeepPhy~\cite{chen2018deepphys}) outperform the traditional methods (e.g., Tulyakov2016~\cite{Tulyakov2016Self}, POS~\cite{wang2017algorithmic}, Haan2013~\cite{de2013robust}) on the challenging VIPL-HR dataset. These results indicate that given a large-scale database (such as VIPL-HR), deep learning based methods are able to learn more informative representation for HR estimation.

We also perform intra-database testing on the MAHNOB-HCI and MMSE-HR databases using a three-fold subject-exclusive protocol as stated in Section~\ref{protocol}. The results can be found in Table~\ref{table:mahnob} and Table~\ref{table:mmse_hr}, respectively. From the results, we can see that the proposed RhythmNet again achieves the best results on both databases. The results indicate that our method has a good generalization ability under different image acquisition conditions, and is fairly robust when the training dataset become small.

\subsubsection{Experiment on NIR Face Video}

\begin{figure}
\centering
\includegraphics[width=0.6\linewidth, height = 0.3\linewidth]{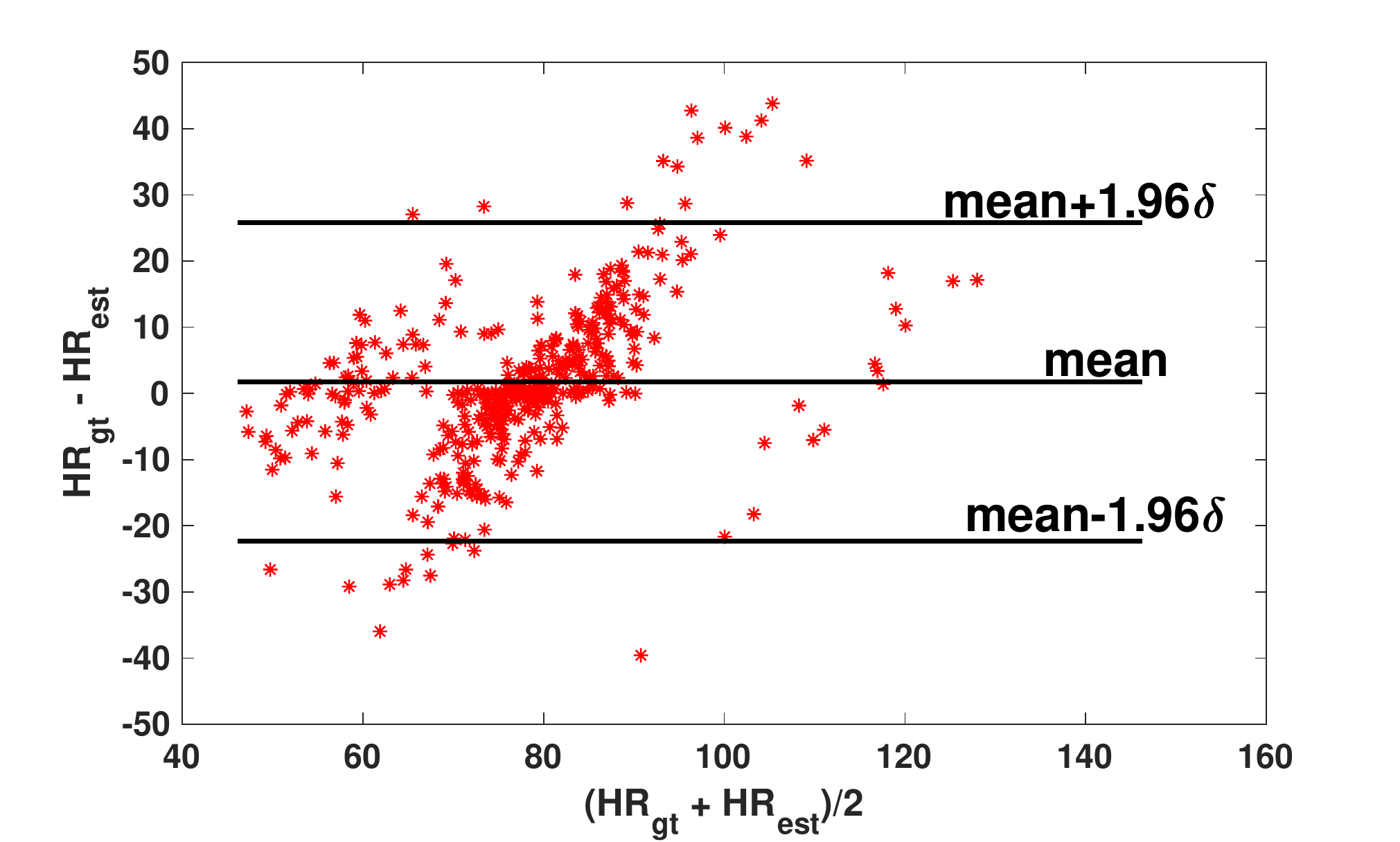}
\caption{ The Bland-Altman plots demonstrating the agreement of the estimated $HR_{est}$ by RhythmNet and the ground-truth $HR_{gt}$ for NIR face videos in VIPL-HR. The lines represent the mean and $95\%$ limits of agreement.}
\label{fig:BAplot_NIR}
\end{figure}

The experiments on NIR face videos are also conducted using the five-fold subject-exclusive protocol. Since the NIR face video only has one channel, we do not use color space transformation and compute one-channel spatial-temporal map for our RhythmNet.
The baseline methods~\cite{de2013robust,Tulyakov2016Self,wang2017algorithmic,chen2018deepphys} for RGB based HR estimation do not work on NIR videos. Thus, we only report the results by our RhythmNet without GRU and the complete RhythmNet in Table.~\ref{table:NIR}. The Bland-Altman plots for NIR face videos is also given in Fig.~\ref{fig:BAplot_NIR}.

\setlength{\tabcolsep}{2pt}
\begin{table}
\begin{center}
\caption{The HR estimation results on NIR face videos of our VIPL-HR database.}
\label{table:NIR}
\begin{tabular}{lcccccc}
\toprule
\multirow{2}{*}{Method} & Mean  & Std  & MAE & RMSE & \multirow{ 2}{*}{MER} & \multirow{ 2}{*}{$r$}\\
&(bpm) &(bpm) &(bpm)&(bpm)\\
\toprule
RhythmNet w/o GRU  & \textbf{1.26} & 13.8 & 9.11 & 13.9 & 11.6\% & 0.62 \\			
RhythmNet & 1.74  & \textbf{12.4} & \textbf{8.45} & \textbf{12.5}  & \textbf{10.8\%}  & \textbf{0.71}\\
\bottomrule
\end{tabular}
\end{center}
\end{table}

From Fig.~\ref{fig:BAplot_color_VIPL} and Fig.~\ref{fig:BAplot_NIR}, we can see that the results using NIR face videos are not as good as those using RGB face videos. This is understandable because a single NIR channel may not convey as sufficient HR rhythm information as the three-channel RGB videos. In addition, we notice the SeetaFace face detector is trained using RGB images, and may fail to detect some faces in the NIR face images.


\subsection{Cross-database Testing}

In order to validate the generalization ability of our RhythmNet, we also perform cross-database evaluations using our VIPL-HR database and two widely used HR estimation databases, i.e., MAHNOB-HCI, and MMSE-HR databases. Specifically, we train our RhythmNet on the color videos of VIPL-HR database and directly test it on the MAHNOB-HCI and MMSE-HR databases. We also fine-tune the VIPL-HR pre-trained model on MAHNOB-HCI and MMSE-HR following a three-fold subject-independent cross-validation testing protocol to see whether per-database fine-tuning could improve the HR estimation performance or not. All the results are provided in Table~\ref{table:mahnob} and Table~\ref{table:mmse_hr}. The baseline methods we use for comparisons are Poh2010~\cite{poh2010non}, Poh2011~\cite{poh2011advancements}, Balakrishnan2013~\cite{balakrishnan2013detecting}, Li2014~\cite{li2014remote}, Haan2013~\cite{de2013robust} and Tulyakov2016~\cite{Tulyakov2016Self}, and we directly use the results of these methods provided in ~\cite{li2014remote} and~\cite{Tulyakov2016Self}.

\setlength{\tabcolsep}{4pt}
\begin{table}
\begin{center}
\caption{The HR estimation results by the proposed approach and several state-of-the-art methods on the MAHNOB-HCI database.}
\label{table:mahnob}
\begin{tabular}{lccccc}
\toprule
\multirow{ 2}{*}{Method} & Mean & Std  & RMSE  & \multirow{ 2}{*}{MER} & \multirow{ 2}{*}{r}\\
&(bpm) &(bpm) &(bpm)&\\
\toprule
Poh2010 \cite{poh2010non}	& -8.95	 &24.3	& 25.9	&25.0\%	& 0.08\\
Poh2011 \cite{poh2011advancements}	& 2.04	&13.5	& 13.6	&13.2\% & 0.36	\\
Balakrishnan2013 \cite{balakrishnan2013detecting}	& -14.4	 &15.2	& 21.0	& 20.7\% & 0.11	\\
Li2014 \cite{li2014remote}	& -3.30	 &6.88	& 7.62	& 6.87\% & 0.81	 \\
Haan2013 \cite{de2013robust} & -2.89 &13.67	& 10.7	&12.9\% & 0.82\\
Tulyakov2016 \cite{Tulyakov2016Self} & 3.19  & 5.81	& 6.23	&5.93\% & 0.83 \\
\hline
\noalign{\smallskip}
RhythmNet(WithinDB) &  \textbf{0.41} &   3.98 &  4.00 &   4.18\%  &   0.87   \\
RhythmNet(CrossDB) &  -5.66  &  6.06   &  8.28 &   8.00\%  &  0.64    \\
RhythmNet(Fine-tuned) & 0.43 &  \textbf{3.97} &   \textbf{3.99} &   \textbf{4.06\%}  &  \textbf{ 0.87 }  \\
\bottomrule
\end{tabular}
\end{center}
\end{table}

\setlength{\tabcolsep}{4pt}
\begin{table}
\begin{center}
\caption{The HR estimation results by the proposed approach and several state-of-the-art methods on the MMSE-HR database.}
\label{table:mmse_hr}
\begin{tabular}{lccccc}
\toprule
\multirow{2}{*}{Method} & Mean  & Std   & RMSE  & \multirow{ 2}{*}{MER} & \multirow{ 2}{*}{$r$}\\
&(bpm) &(bpm) &(bpm)&\\
\toprule
Li2014~\cite{li2014remote} & 11.56 & 20.02 & 19.95 & 14.64\% & 0.38 \\
Haan2013~\cite{de2013robust} & 9.41 & 14.08 & 13.97 & 12.22\% & 0.55 \\
Tulyakov2016~\cite{Tulyakov2016Self} & 7.61 &12.24 & 11.37 & 10.84\% & 0.71 \\
\hline
\noalign{\smallskip}
RhythmNet(WithinDB) & \textbf{-0.85} &  \textbf{4.99} &  \textbf{5.03} &   3.67\%  &  \textbf{0.86}   \\
RhythmNet(CrossDB) & -.2.33 &  6.98 &  7.33 &   3.62\%  &   0.78   \\
RhythmNet(Fine-tuned) & -0.88 &   5.45  & 5.49 &  \textbf{3.58\%} & 0.84 \\
\bottomrule
\end{tabular}
\end{center}
\end{table}

From the results, we can see that the proposed method could achieve promising results even when we directly test our VIPL-HR pre-trained model on these two databases, i.e., a RMSE of 8.28 bpm on the MAHNOB-HCI database, and a RMSE of 7.33 bpm on the MMSE-HR database. The error rates are further reduced to 3.99 bpm and 5.49 bpm on MAHNOB-HCI and MMSE-HR, respectively when we fine-tune the pre-trained model on MAHNOB-HCI and MMSE-HR databases. Both results by the proposed approach are much better than previous methods. These results indicate that the variations of illumination, movement, and acquisition sensors covered in the VIPL-HR database are helpful for learning a deep HR estimator which has good generalization ability to unseen scenarios. In addition, the proposed RhythmNet is able to leverage the diverse image acquisition conditions in VIPL-HR to learn a robust HR estimator.

\subsection{Key Components Analysis}
\label{Key_components_analysis}

In this section, we further analyze the effectiveness of our approach from multiple aspects, covering spatial-temporal map, color space selection, temporal modeling via GRU and computational cost. All the experiments are conducted on the VIPL-HR database under a subject-exclusive five-fold cross-validation protocol.

\subsubsection{Spatial-temporal Map}
\label{spatial-temporal map}

\begin{figure}
\centering
\subfigure[I3D]{
\includegraphics[width=0.22\linewidth, height=0.22\linewidth]{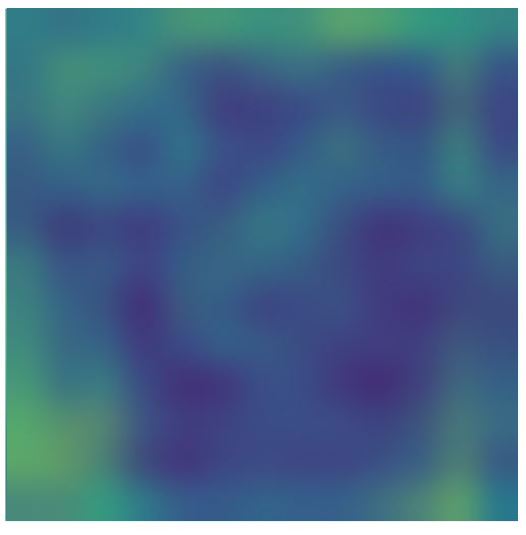}\vspace{8pt}
\label{fig:mean_vis_I3D}
}
\subfigure[RhythmNet]{
\includegraphics[width=0.264\linewidth, height=0.22\linewidth]{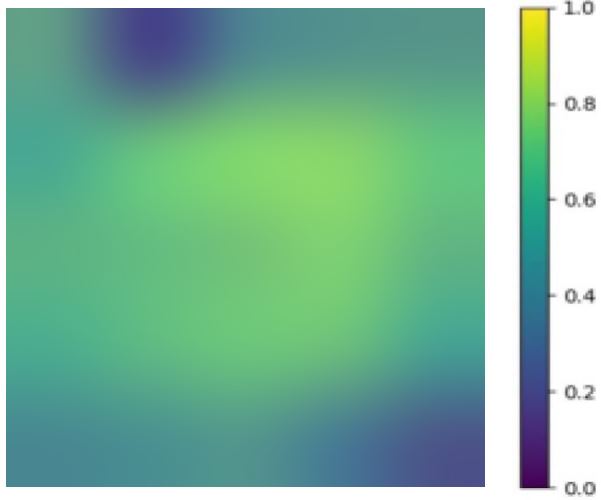}
\label{fig:mean_vis_STmap}
}
\caption{ The mean visualization maps of all the video clips in the test set of one fold for (a) I3D model~\cite{carreira2017quo} and (b) our RhythmNet.}
\label{fig:I3D_visualization}
\end{figure}

From the results in Table~\ref{table:vipl_color}, we can see that under the same experimental settings, I3D~\cite{carreira2017quo} trained from raw RGB face videos achieves a RMSE of 15.9 bpm, which is much worse than our RhythmNet with spatial-temporal maps as representations (RMSE = 8.14 bpm). These results indicate that it is very difficult for deep neural networks to directly learn informative features from the raw RGB videos.
We further visualize the feature maps of I3D for the input video clips using the visualization method proposed in~\cite{yu2019remote}. The visualization map is calculated based on the average of all the feature maps of all video clips, i.e., 16427 video clips from 476 videos in one fold of test. The map is shown in Fig.~\ref{fig:mean_vis_I3D}. The visualization map of our RhythmNet is given in Fig.~\ref{fig:mean_vis_STmap}. Since our RhythmNet takes the spatial-temporal maps as input, we calculated the mean responses of each rows (spatial dimension) and then reshaped the responses to a visualization map. From the visualization maps, we can see that our RhythmNet focuses on the informative face regions, and thus gives more accurate estimation results.

\subsubsection{Color Space Selection}
\label{color_space}
\begin{figure}
\centering
\includegraphics[width=0.6\linewidth]{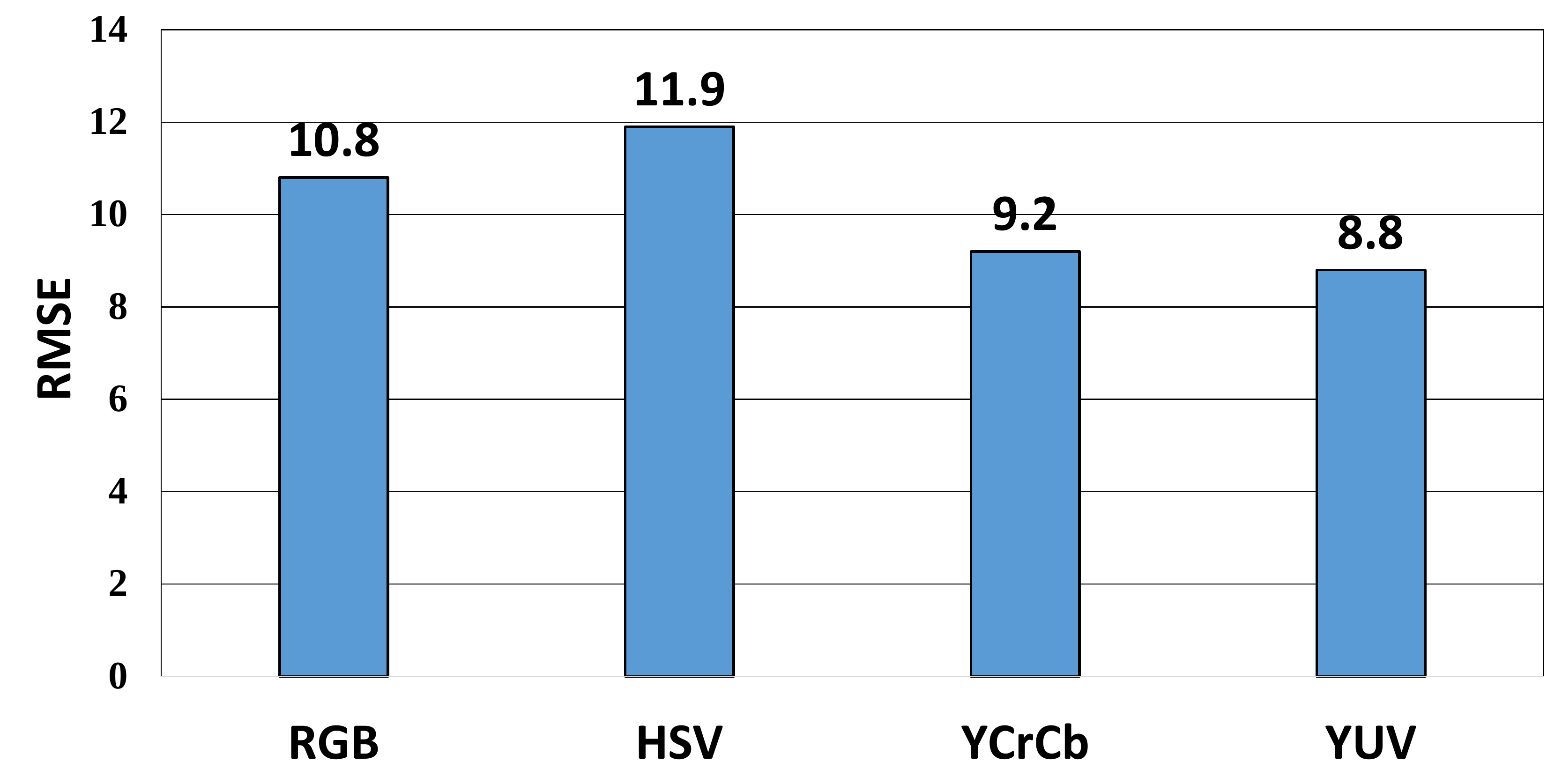}
\caption{ The HR estimation errors (in RMSE) by RhythmNet using spatial-temporal maps generated from different color spaces (RGB, HSV, YCrCb, and YUV). }
\label{fig:color_space}
\end{figure}

As discussed in~\cite{yang2016motion}, head movement effects much more on the intensity than the chromaticity of the image since the chromaticity reflects the intrinsic optical properties of hemoglobin in blood. Therefore, choosing a color space separating chromaticity from intensity is helpful to reduce the artifacts and thus benefit the training procedure. We considered three commonly used color space transformations, i.e., HSV, YCrCb, and YUV, as well as the RGB color space. For each experiment, the color space used for testing is the same as the color space used for training. The results by RhythmNet using these color spaces are given in Fig.~\ref{fig:color_space}.

From the results, we can see that different color spaces lead to very different HR estimation accuracies, and using YUV color space achieves the best result with a RMSE of 8.8 bpm. The results indicate that the YUV color space is more suitable in representing the HR signals, and helpful for capturing more information about heart rhythm. This is the reason that we choose to use the YUV color space in all the experiments.

\subsubsection{Temporal Modeling}

For HR measurement, we use GRU to model the relationship between adjacent measurements from succeeding video clips.
We study the importance of this part by removing it from our RhythmNet and direct compute the average HR of estimations for all the clips. The results by RhythmNet with and without using the GRU are given in Table~\ref{table:vipl_color}. From the results, we can see that removing the GRU module leads to increased RMSE from 8.11bmp to 8.94 bpm, which suggests the usefulness of the temporal modelling module in our RhythemNet.

\subsubsection{Computational Cost}

The model size of our HR estimator is about 42 MB, and the inference time is about 8ms on a Titan 1080Ti GPU when using a sliding window with 300 frames. These results indicate that the proposed approach is efficient, enabling real-time HR measurement.

\subsection{Further Analysis}
\label{further_analysis}

While the proposed RhythmNet achieves promising results on multiple databases, robust remote HR estimation under less-constrained scenarios faces a number of challenges. We provide additional analysis in terms of the challenging factors, i.e., video compression, illumination, head movement.

\subsubsection{Video Compression}
\label{video_compression}

As shown in Section~\ref{compression}, when using MJPG 
to compress the video data, the state-of-the-art method~\cite{de2013robust} shows minor performance degradation. We are also interested to see how such a compression can influence the proposed RyhthmNet. As shown in Table~\ref{table:compression}, the HR estimation errors by our RhythmNet before and after compression are also the same. This suggests that the proposed approach is also very robust to the video compression by MJPG. From the result, we can see that the estimated results based on compressed data are very close to the results based on uncompressed data, which indicates that the compressed version VIPL-HR database is able to maintain the HR signals. All the experiments in this paper are based on the compressed version of VIPL-HR database.

\setlength{\tabcolsep}{4pt}
\begin{table}
\begin{center}
\caption{The HR estimation results by our RyhthmNet using compressed and uncompressed RGB face videos of VIPL-HR.}
\label{table:compression}
\begin{tabular}{lcccccc}
\toprule
\multirow{2}{*}{Datatype} & Mean  & Std  & MAE & RMSE  & \multirow{ 2}{*}{MER} & \multirow{ 2}{*}{$r$}\\
&(bpm) &(bpm) &(bpm)&(bpm)\\
\toprule
Compressed & 0.34 & 8.76  & 5.94 & 8.77 & 7.54\% & 0.58 \\
Uncompressed & 0.08 &  8.69 &  5.81 &  8.69 &  7.24\%  & 0.59 \\
\bottomrule
\end{tabular}
\end{center}
\end{table}

\subsubsection{Illumination}
\begin{figure}
\centering
\subfigure[]{
\includegraphics[width=0.42\linewidth]{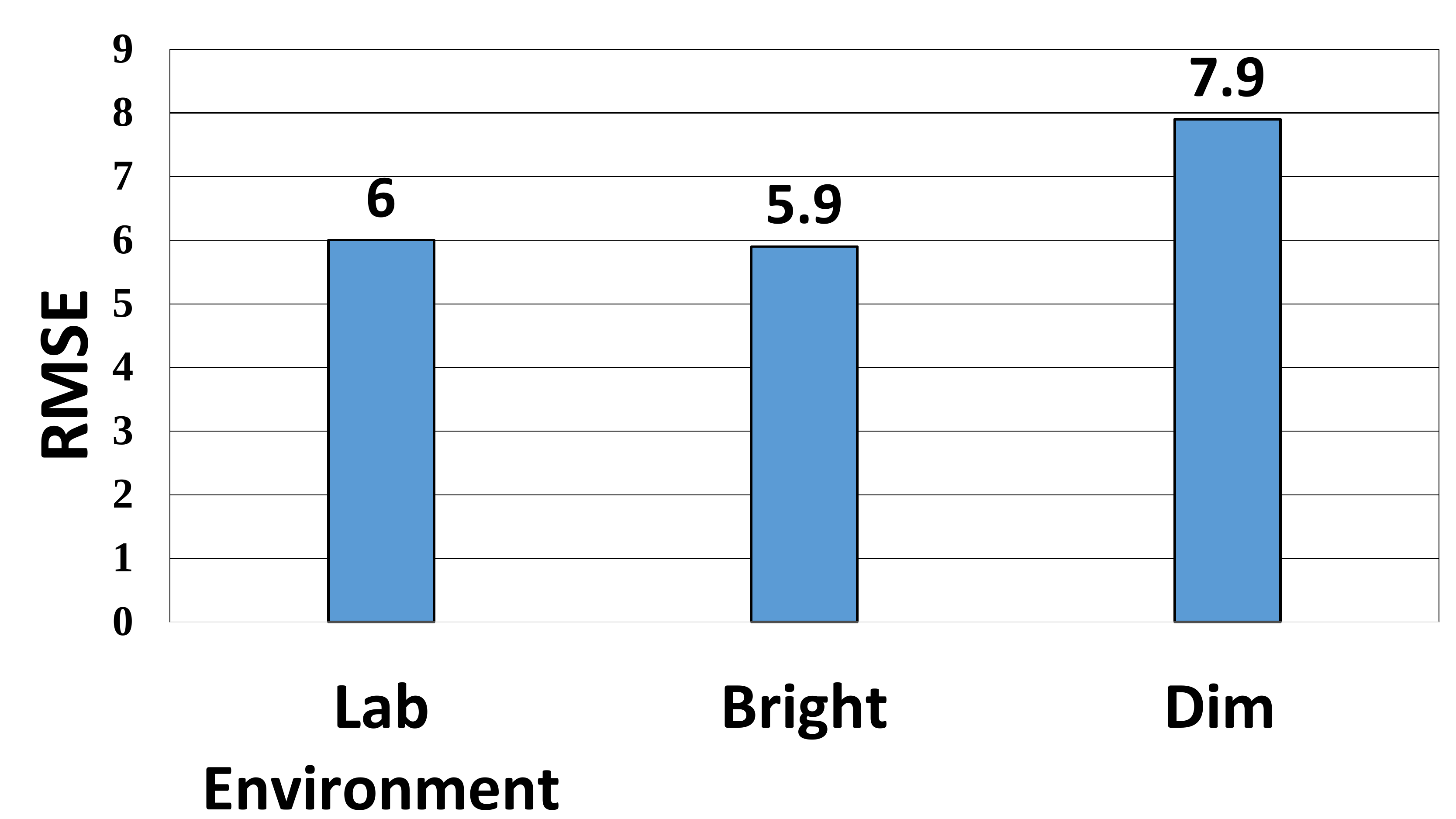}
\label{fig:analysis_illumination}
}
\subfigure[]{
\includegraphics[width=0.42\linewidth]{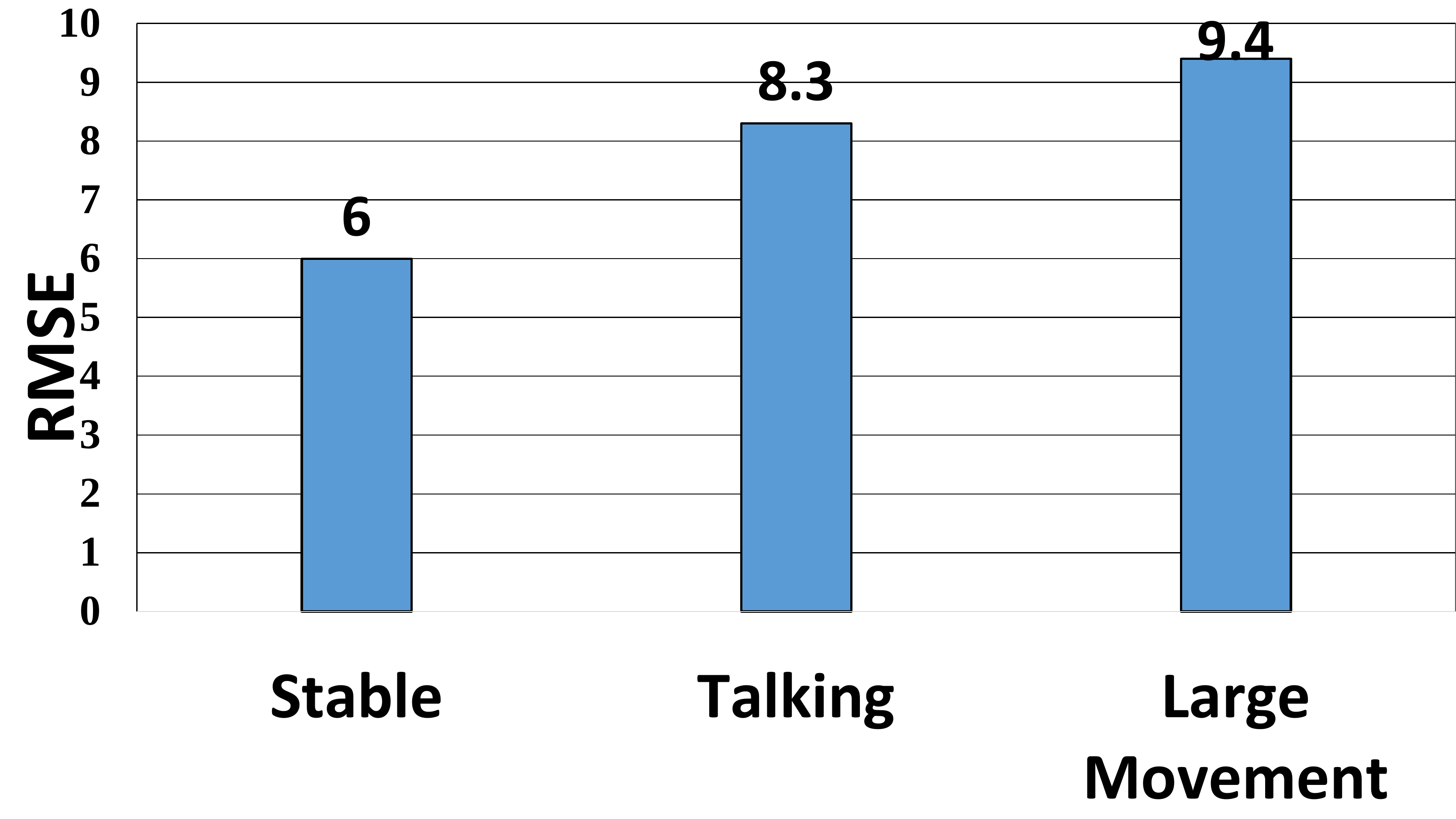}
\label{fig:analysis_motion}
}
\caption{ The HR estimation errors (in RMSE) by RhythmNet under (a) different illumination conditions and (b) different head movement conditions. }
\end{figure}

We study the influence of illumination variations on our RhythemNet by using the RGB face videos of Scenario 1, 4, and 5 in Table~\ref{table:record_situation} from the VIPL-HR dataset. The RMSE are reported in Fig.~\ref{fig:analysis_illumination}. From the results, we can see that under a bright lighting condition, i.e., with the lamp placed in front of the subject, our RhythemNet achieves a lower RMSE of 5.90 bpm comparing with the result with lab environment illumination. At the same time, under dim light conditions, the estimation error raises to 7.94 bpm. These results indicate that a good light condition is important for improving the HR estimation accuracy.
However, even under dim light condition, the HR estimation error by our RhythmNet is lower than 8 bpm, which an encouraging result for practical application usage.

\begin{figure}
\centering
\subfigure[]{
\includegraphics[width=0.42\linewidth]{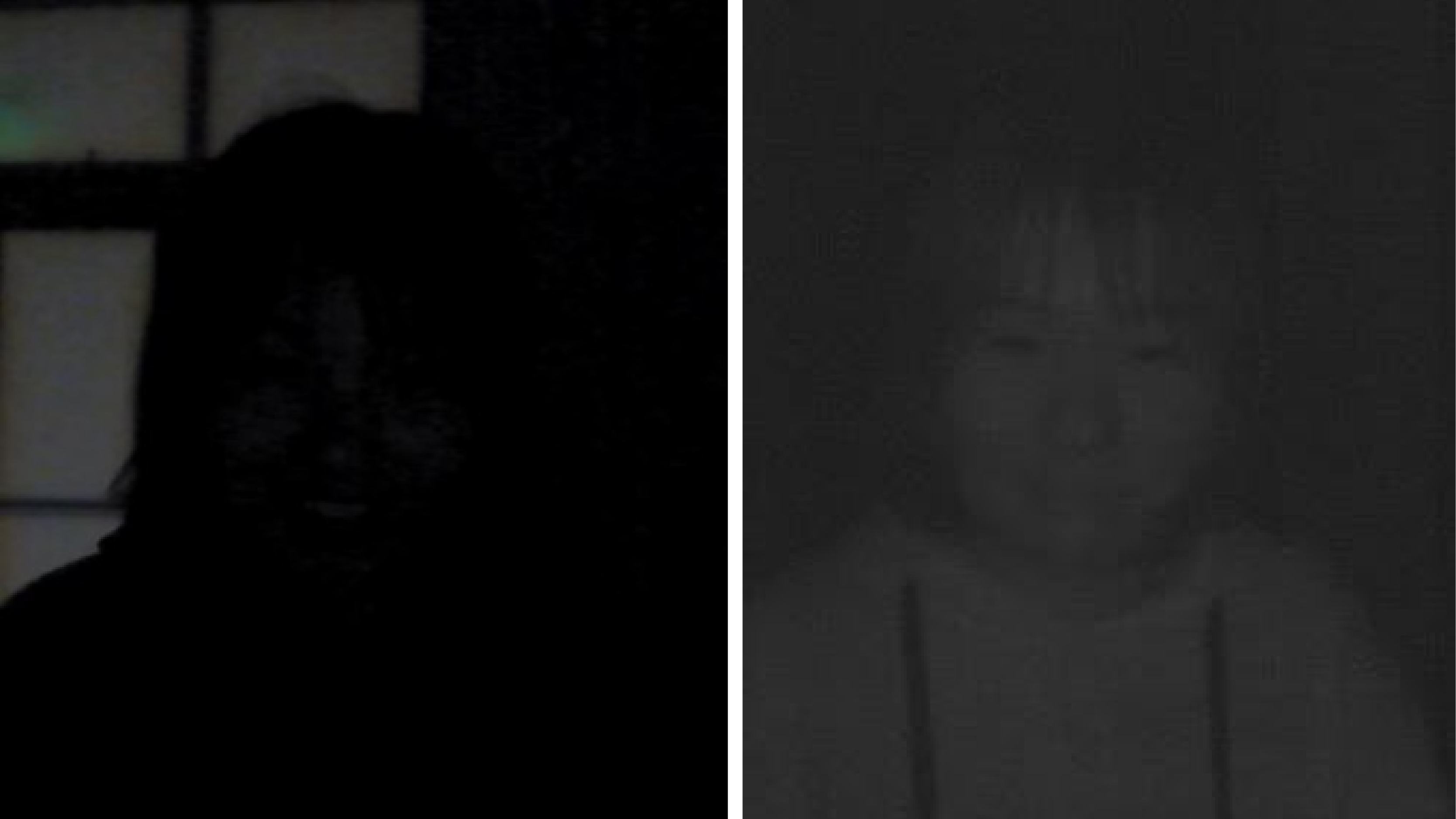}
\label{fig:dark}
}
\subfigure[]{
\includegraphics[width=0.42\linewidth]{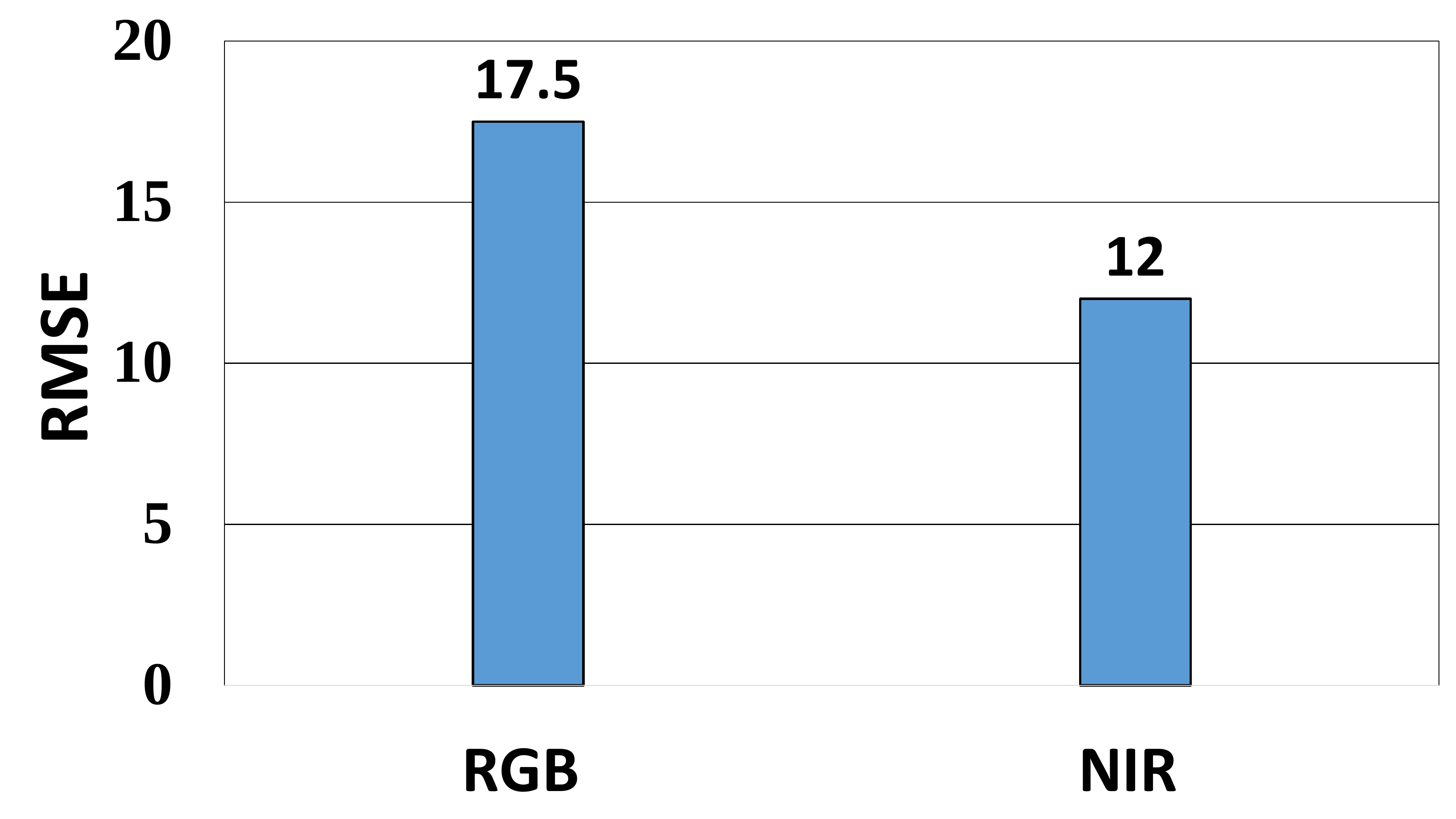}
\label{fig:dark_result}
}
\caption{ (a) Examples of a RGB (left) and a NIR (right) face videos captured in dark scenario. (b) The corresponding HR estimation errors (in RMSE) by RhythmNet trained on RGB and NIR videos.  }
\end{figure}

We also study the influence of RGB and NIR videos for HR estimation under dark scenario by user tests with 6 subjects. As shown in Fig.~\ref{fig:dark}, we have recorded 8 RGB videos and 8 NIR videos for testing. The RhythmNet used for testing remains the one trained on the VIPL-HR. The HR estimation results using RGB and NIR videos are given in Fig.~\ref{fig:dark_result}. As expected, we can see that under dark environment, RhythemNet gets an increased HR estimation error than under bright environment (from 8.14 bmp in RMSE to 17.7 bpm in RMSE on VIPL-HR). Using NIR videos for HR estimation can reduce the HR estimation error to RMSE = 12 bpm under the same dark environment. These results indicate that NIR face videos can be a good choice for HR measurement in dark environment.

\subsubsection{Head Movement}

In order to analyse the influence of head movements, we compute the HR estimation errors for Situation 1, 2, and 3 in Table~\ref{table:record_situation} from the VIPL-HR dataset. The results are shown in Fig.~\ref{fig:analysis_motion}. From the results, we can see that all kinds of head movements could significantly influence the estimation accuracy. The RMSE raises to 8.28 bpm when the subject is talking, and 9.40 bpm when the subject performs large scale head movements. These results indicate that HR estimation under different head movements is still a challenging task, and should be considered when building a robust HR estimator.


\section{Conclusion and Future Work}
\label{conculsion}

While remote HR estimation from a face video has wide applications, it is a challenging problem in less-constrained scenarios due to variations of head movement, illumination, and sensor diversity. In this paper, we propose an end-to-end learning approach (RhythmNet) for robust heart rate estimation from face videos. Our RhythmNet uses a novel spatial-temporal map represent the HR signals in the videos, which allows efficient modelling via CNN. We also take into account the temporal relationship of adjacent HR estimates to perform continuous HR measurement. At the same time, we also build a multi-modality database (VIPL-HR) for studying HR estimation under less-constrained scenarios covering variations of illumination, head movement, and sensor diversity. The proposed approach achieves promising HR estimation accuracies in both within-database and cross-database testing scenarios on the MAHNOB-HCI, MMSE-HR, and VIPL-HR databases. In our future work, we would like to investigate the effectiveness of the proposed approach for the other rPPG-based physiological status measurement tasks, such as the breath rate and blood pressure measurement from videos. We would also like to investigate robust HR estimation approaches leveraging distribution learning~\cite{pan2018mean-variance} or multi-task learning~\cite{wang2017deep,cui2018improving}.

\section*{Acknowledgment}
The authors would like to thank the editors and anonymous reviewers for their constructive comments and suggestions.
\ifCLASSOPTIONcaptionsoff
  \newpage
\fi

\bibliographystyle{IEEEtran}
\bibliography{egbib}

\begin{IEEEbiography}[{\includegraphics[width=1in,height=1.25in,clip,keepaspectratio]{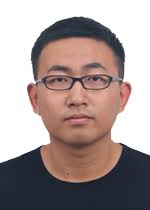}}]{Xuesong Niu} received the B.E. degree from Nankai University and is pursing the Ph.D. degree from Institute of Computing Technology (ICT), Chinese Academy of Sciences (CAS), Beijing, China. His research interests include computer vision, machine learning and affective computing.
\end{IEEEbiography}

\begin{IEEEbiography}[{\includegraphics[width=1in,height=1.25in,clip,keepaspectratio]{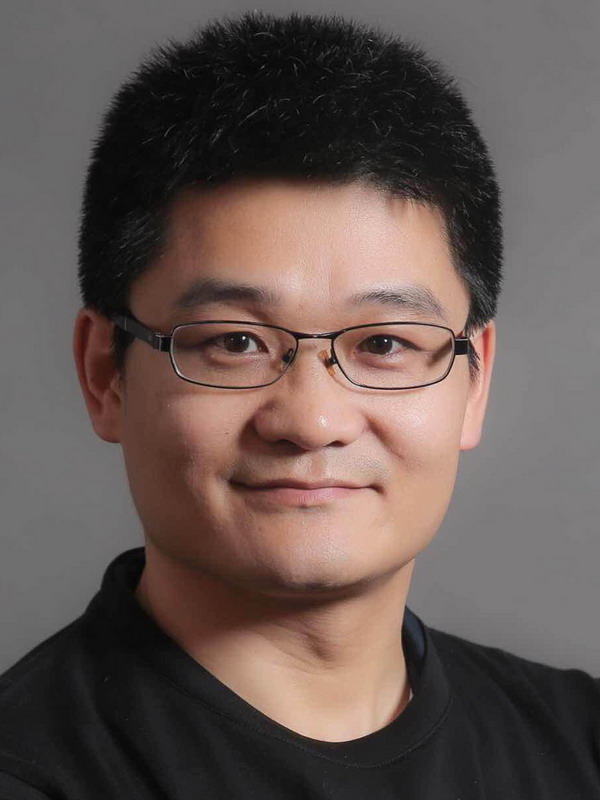}}]{Shiguang Shan} (M'04-SM'15)
received Ph.D. degree in computer science from the Institute of Computing Technology (ICT), Chinese Academy of Sciences (CAS), Beijing, China, in 2004. He has been a full Professor of this institute since 2010 and now the deputy director of CAS Key Lab of Intelligent Information Processing. He is also a member of CAS Center for Excellence in Brain Science and Intelligence Technology. His research interests cover computer vision, pattern recognition, and machine learning. He has published more than 300 papers, with totally more than 15,000 Google scholar citations. He served as Area Chairs for many international conferences including ICCV11, ICASSP14, ICPR12/14/19, ACCV12/16/18, FG13/18/20, BTAS18 and CVPR19/20. And he was/is Associate Editors of several journals including IEEE T-IP, Neurocomputing, CVIU, and PRL. He was a recipient of the China's State Natural Science Award in 2015, and the China's State S\&T Progress Award in 2005 for his research work.
\end{IEEEbiography}

\begin{IEEEbiography}[{\includegraphics[width=1in,height=1.25in,clip,keepaspectratio]{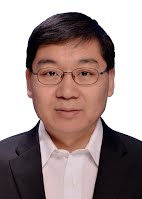}}]{Hu Han} is an Associate Professor of the Institute of Computing Technology (ICT), Chinese Academy of Sciences (CAS). He received the B.S. degree from Shandong University, and the Ph.D. degree from ICT, CAS, in 2005 and 2011, respectively, both in computer science. Before joining the faculty at ICT, CAS in 2015, he has been a Research Associate at PRIP lab in the Department of Computer Science and Engineering at Michigan State University, and a Visiting Researcher at Google in Mountain View. His research interests include computer vision, pattern recognition, and image processing, with applications to biometrics and medical image analysis. He has authored or co-authored over 50 papers in refereed journals and conferences including IEEE Trans. PAMI/IFS, CVPR, ECCV, NeurIPS, and MICCAI. He was a recipient of the IEEE FG2019 Best Poster Presentation Award, and CCBR 2016/2018 Best Student/Poster Awards. He is a member of the IEEE.
\end{IEEEbiography}

\begin{IEEEbiography}[{\includegraphics[width=1in,height=1.25in,clip,keepaspectratio]{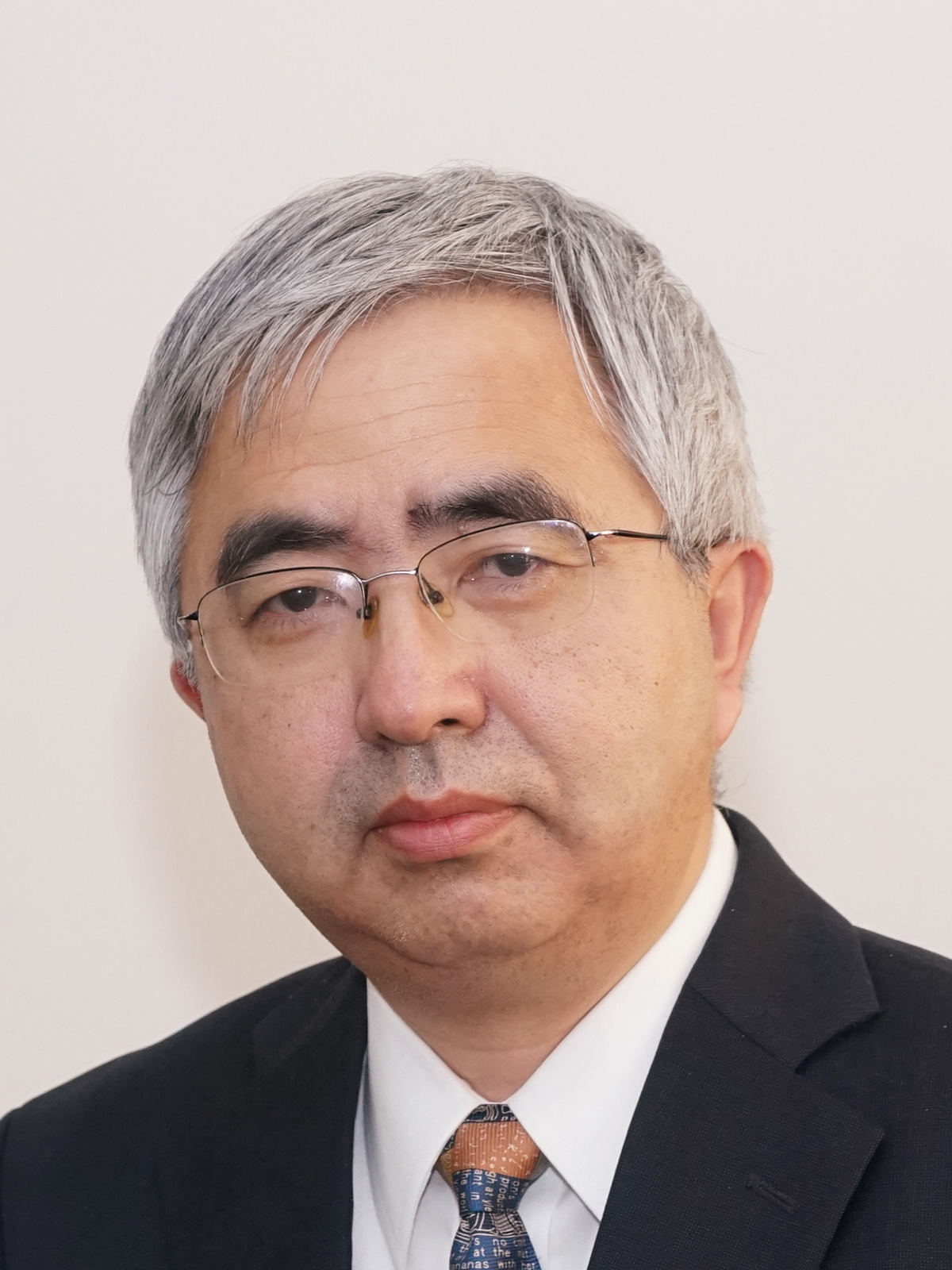}}]{Xilin Chen} is a professor with the Institute of Computing Technology, Chinese Academy of Sciences (CAS). He has authored one book and more than 300 papers in refereed journals and proceedings in the areas of computer vision, pattern recognition, image processing, and multimodal interfaces. He is currently an associate editor of the IEEE Transactions on Multimedia, and a Senior Editor of the Journal of Visual Communication and Image Representation, a leading editor of the Journal of Computer Science and Technology, and an associate editor-in-chief of the Chinese Journal of Computers, and Chinese Journal of Pattern Recognition and Artificial Intelligence. He served as an Organizing Committee member for many conferences, including general co-chair of FG13 / FG18, program co-chair of ICMI 2010. He is / was an area chair of CVPR 2017 / 2019 / 2020, and ICCV 2019. He is a fellow of the IEEE, IAPR, and CCF.
\end{IEEEbiography}

\end{document}